\documentclass[lettersize,journal]{IEEEtran}
\usepackage{amsmath,amsfonts}
\usepackage{algorithmic}
\usepackage{algorithm}
\usepackage{array}
\usepackage[caption=false,font=scriptsize,labelfont=sf,textfont=sf]{subfig}
\usepackage{textcomp}
\usepackage{stfloats}
\usepackage{url}
\usepackage{hyperref}
\usepackage{verbatim}
\usepackage{graphicx}
\usepackage{cite}
\usepackage{amsmath}    %
\usepackage{amssymb}
\usepackage{amsthm}
\usepackage{amsfonts}
\usepackage{stackrel}  
\usepackage{dsfont}
\usepackage{multirow}
\usepackage{color}
\usepackage{booktabs, tabularx}
\usepackage{makecell}
\newtheorem{definition}{Definition}

\newtheorem{remark}{Remark}

\renewcommand{\algorithmicrequire}{\textbf{Input:}}
\renewcommand{\algorithmicensure}{\textbf{Output:}}

\begin{document}

\title{Graph Memory Learning: Imitating Lifelong Remembering and Forgetting of Brain Networks}

\author{Jiaxing Miao, Liang Hu,~\IEEEmembership{Member~IEEE}, Qi Zhang, and Longbing Cao,~\IEEEmembership{Senior Member~IEEE}
\thanks{Jiaxing Miao is  with the Department of Computer Science and Technology, Tongji University, Shanghai 201804, China (e-mail: JiaxingMiao@tongji.edu.cn)}
\thanks{Liang Hu and Qi Zhang are with the Department of Computer Science and Technology, Tongji University, Shanghai 201804, China (e-mail: {zhangqi\_cs,
lianghu}@tongji.edu.cn).}
\thanks{Longbing Cao is with the School of Computing, Macquarie University,
Sydney, NSW 2109, Australia (e-mail: longbing.cao@mq.edu.au).}
\thanks{Manuscript received April 19, 2021; revised August 16, 2021. 
\emph{(Corresponding author: Liang Hu.)}}}

\markboth{Journal of \LaTeX\ Class Files,~Vol.~14, No.~8, August~2021}%
{Shell \MakeLowercase{\textit{et al.}}: A Sample paper Using IEEEtran.cls for IEEE Journals}


\maketitle

\begin{abstract}
Graph data in real-world scenarios undergo rapid and frequent changes, making it challenging for existing graph models to effectively handle the continuous influx of new data and accommodate data withdrawal requests. The approach to frequently retraining graph models is resource intensive and impractical. To address this pressing challenge, this paper introduces a new concept of \textit{graph memory learning}. Its core idea is to enable a graph model to selectively remember new knowledge but forget old knowledge. Building on this approach, the paper presents a novel graph memory learning framework - \textbf{B}rain-inspired \textbf{G}raph \textbf{M}emory \textbf{L}earning (BGML), inspired by brain network dynamics and function-structure coupling strategies. BGML incorporates a multi-granular hierarchical progressive learning mechanism rooted in feature graph grain learning to mitigate potential conflict between memorization and forgetting in graph memory learning. This mechanism allows for a comprehensive and multi-level perception of local details within evolving graphs. In addition, to tackle the issue of unreliable structures in newly added incremental information, the paper introduces an information self-assessment ownership mechanism. This mechanism not only facilitates the propagation of incremental information within the model but also effectively preserves the integrity of past experiences. We design five types of graph memory learning tasks: regular, memory, unlearning, data-incremental, and class-incremental to evaluate BGML. Its excellent performance is confirmed through extensive experiments on multiple node classification datasets.
\end{abstract}


\section{Introduction}\label{chap:1}
\begin{figure}[t]
	\centering
	\includegraphics[width=\linewidth]{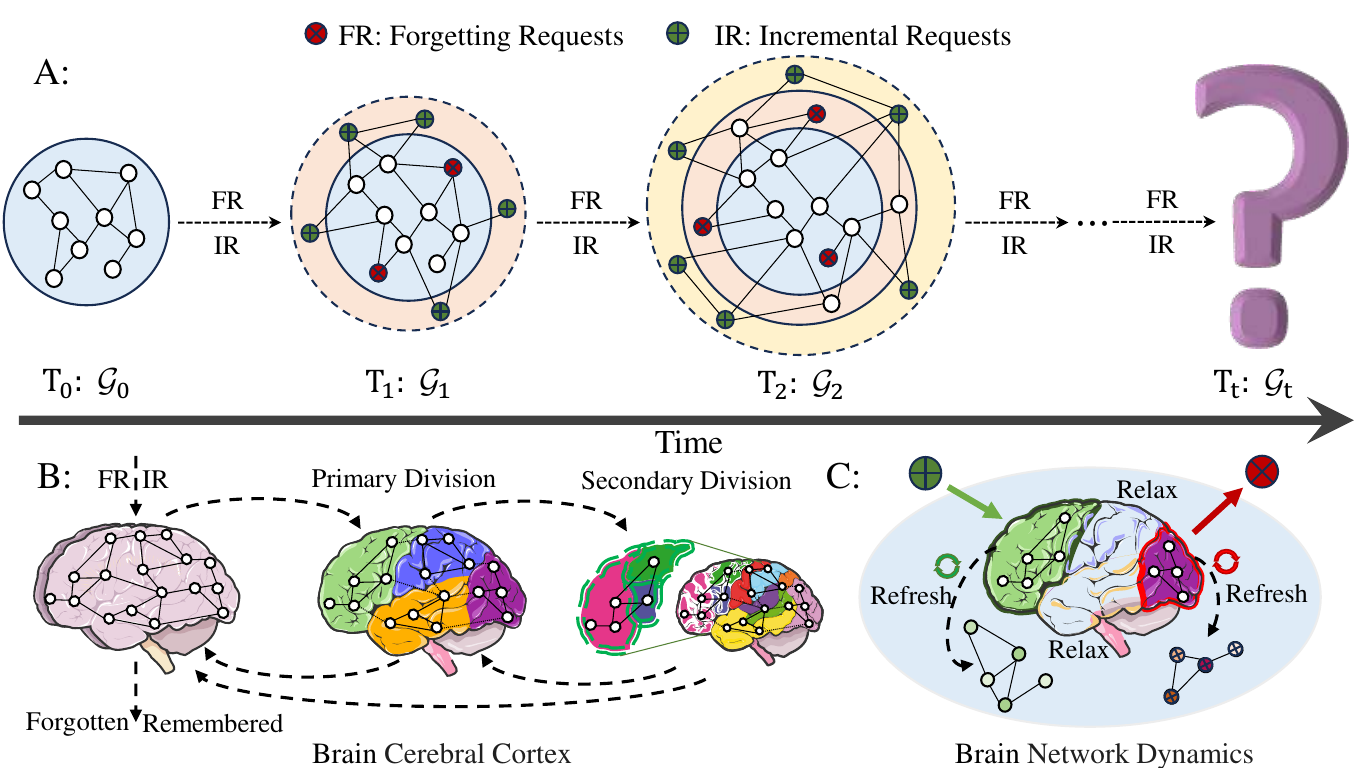}
	\caption{Insights and reflections provided by brain network dynamics for GML. The modular and hierarchical structures of the cerebral cortex achieve function-structure decoupling, which can help the brain adapt to the evolution of knowledge at any time. The remembering and forgetting of knowledge is dynamic because it directly stimulates only the most relevant neurons. \textbf{A:} Graph Evolution in Graph Memory Learning. \textbf{B:} Modular and Hierarchical Structures based on Brain Cerebral Cortex. \textbf{C:} A Dynamic Integration Mechanism based on Brain Network Dynamics.} 
	\label{fig:Graph Evolution}
\end{figure}

\IEEEPARstart{N}{owadays}, a multitude of pivotal datasets in practical applications are expressed as graphs \cite{wu2020comprehensive}, encompassing diverse domains such as social networks \cite{peng2022reinforced, fan2022social}, traffic networks \cite{YiZFHHWACN23, Zheng2024Spatio}, biological networks \cite{ma2023single, Yang2024Inte}, recommendation systems \cite{wu2023generic}. Graph learning models attract increasing attention and demonstrate robust data mining and representation learning capacities. However, graph learning faces the challenge of promptly assimilating new information while discarding obsolete knowledge that has been invalidated. This is attributed to the fact that real-world data often exhibits dynamic flow characteristics, wherein a continuous influx and withdrawal of data occur. As depicted in Fig. \ref{fig:Graph Evolution} (A), the flow of data across the graph frequently induces alterations in the overarching topology, thereby exacerbating the difficulties encountered by graph models. Consequently, the objective of this study is to explore and empower graph learning models with the capability to ``memory" based on the dynamic graph. This paradigm is named as \textit{graph memory learning} (GML).

Existing methods in Graph Lifelong Learning (GLL) \cite{febrinanto2023graph, tian2024continual} and Graph Unlearning (GUL) \cite{said2023survey, wu2023gif} explore memory learning in graph models from a limited perspective. GLL mainly addresses the challenge of continuous data increments, aiming to adapt to new knowledge while mitigating the forgetting of past experiences \cite{wang2022lifelong, zhang2022hierarchical}. For instance, PI-GNN proposed by Zhang et al. \cite{zhang2023continual} handles data increments through parameter isolation and expansion, while Su et al. \cite{su2023towards} tackle catastrophic forgetting through structural alignment. However, GLL faces significant limitations when dealing with dynamic changes in graph data streams within the GML problem. Specifically, GLL fails to effectively adapt to significant changes in the local graph structure when data needs to be retracted. On the other hand, GUL emphasizes the right to forget data, aiming to forget specified retracted data at the lowest possible cost while minimizing harm to existing knowledge \cite{chen2022graph}. For example, the certified unlearning method proposed by Chien et al. \cite{chien2022certified} regularizes data retraction using convex function properties. Methods like IDEA \cite{dong2024idea} and ScaleGUN \cite{yis2025calable} have made improvements in flexibility and scalability. However, these methods primarily focus on static graph data and do not address dynamic graph learning scenarios within the GML problem, nor can they adapt to changes in the new knowledge environment based on forgetting. More deeply, GML is not simply a combination of GLL and GUL. It requires not only solving the challenges faced by them, but also a comprehensive and meticulous consideration of memory management in the context of continuous GML data flow. This is because \textbf{the inherent conflict between memory retention and forgetting} forms a natural barrier, preventing graph learning models from accurately perceiving changes in graph structure, thus leading to a decline in their expressive power. This conflict \textbf{represents the most prominent contradiction in realizing the GML paradigm}.

Upon a deeper examination of the conflict between memory retention and forgetting, we identify two fundamental challenges to be effectively explored in GML. \textbf{Firstly}, the influx and withdrawal of graph data are often closely linked to the node and edge knowledge within the local graph structure. However, frequent changes in this structure prevent graph models from clearly distinguishing between retained and forgotten knowledge. \textbf{Secondly}, retaining emerging information is intricate and unreliable in the graph structure. Specifically, new information can be accidentally and inappropriately linked to irrelevant graph regions. Furthermore, the worst is that new knowledge with significant disparities can unintentionally fill the structural gaps left by the removal of old knowledge. It potentially suppresses knowledge retention and exacerbates catastrophic forgetting. These conditions inhibit the propagation of new knowledge and exacerbate catastrophic forgetting.

Fortunately, the function-structure coupling brain can be considered a remarkable example of "natural" memory intelligence. It possesses the capacity to absorb new knowledge and integrate old knowledge continually through structural brain networks. When confronted with disruptive stimuli, the brain can even selectively remodel synapses and forget pertinent memories as a protective measure \cite{eichenbaum2004conditioning}. Intuitively, network neuroscience holds inspiring insights into addressing the above challenges in GML. \textbf{Firstly}, the brain employs a modular and hierarchical structure (Fig. \ref{fig:Graph Evolution} (B)), enabling it to perceive local changes with precision and distinguish between remembered and forgotten information. This facilitates the comprehension of evolving trends in knowledge alteration~\cite{eichenbaum2007medial}. \textbf{Secondly}, the brain exhibits a dynamic integration mechanism (Fig. \ref{fig:Graph Evolution} (C)) to adjust brain networks based on dynamics \cite{preston2013interplay}. Modifications in local knowledge trigger responses exclusively in corresponding brain regions, ensuring efficient recognition of changes with minimal energy expenditure while preserving past experiences in inactive brain regions. This mechanism directs the flow of knowledge to the designated brain regions based on relevance and activates neurons that store the most pertinent old experiences for integration. It precisely helps the brain minimize interference from previous experiences.

Therefore, by drawing insights from the modular and hierarchical structure of the brain, the graph can be divided into multi-level subgraphs based on node attributes and structural characteristics. This progressive and multi-granular approach to learning graph knowledge enables graph models to distinguish the boundaries of knowledge retention and forgetting. For instance, when a recommendation system learns preferences of different granularities by building labels on various levels, the system obtains a robust and deep cognition of user nodes, and the influence of alterations in neighboring nodes on recommendations diminishes. Additionally, a dynamic integration system for new knowledge can be designed, akin to the brain network dynamics. The system employs the $L_{2}$ distance to guide new knowledge toward credible a local graph structure and integrate it with relevant past experiences. For instance, in a recommendation system, when new users form unreliable connections due to accidental purchases, recommendations based on the most similar users are the most reliable and mitigate the impact of old experiences. Furthermore, removing the constraints imposed by the graph structure facilitates the implementation of the above concepts within GML. Ideally, nodes should function as independent samples capable of efficient and scalable batch processing.

In light of the above observations, this paper proposes a \textbf{B}rain-inspired \textbf{G}raph \textbf{M}emory \textbf{L}earning algorithm (\textbf{BGML}) based on brain network dynamics and function-structure coupling. It aims to solve the problem of GML for dynamically evolving graph data. Specifically, this paper proposes a multi-granular feature graph learning scheme with hierarchical progressive learning to clarify changes in local graph knowledge. Key components comprise a multi-granular graph partition (MGP) and a multi-granular hierarchical progressive learning mechanism (MGHPL) based on progress-aware module. Notably, progress-aware module evaluates and provides real-time feedback on the cognitive capacity of each sub-model, dynamically adjusting the integrated framework. Furthermore, an information self-assessment ownership mechanism (ISAO) is proposed to tackle the issue of unreliable new knowledge structures. Essentially, this paper explores three questions: \textbf{\emph{(1) What GML approach can empower graph models to selectively remember important information and forget irrelevant information? (2) How to alleviate the conflict between remembering and forgetting in the graph memory model? (3) How to assign a credible graph structure to knowledge during the process of remembering and forgetting?}} The main contributions are summarized below:
\begin{itemize}
\item This paper makes the first attempt to define a new problem of Graph Memory Learning and to propose a novel GML paradigm inspired by brain network dynamics and function-structure coupling. 
\item A multi-granular hierarchical progressive learning mechanism based on feature graph grain learning enables GML to alleviate the conflict between remembering and forgetting in local graph structures.
\item We propose an information self-assessment ownership mechanism to establish a reliable graph structure for new knowledge such that new knowledge can independently choose the most suitable graph grain and neighbors.
\item Extensive experiments on five tasks targeting different aspects of GML on the node classification task on nine datasets demonstrate that BGML outperforms all baselines significantly.
\end{itemize}

\section{Related Work}\label{chap:2}
\subsection{Graph Neural Networks}
Graph neural networks (GNNs) have become a prominent method for processing unstructured data \cite{li2023graph, han2024bigst, wu2022graph}. Specifically, Kpif et al. \cite{TKipf2017} introduced GCNs, utilizing first-order truncated Chebyshev polynomials to approximate graph convolutions. Additionally, Veličković et al. \cite{velivckovicgraph2018} introduced Graph Attention Networks (GATs), incorporating an attention mechanism on the graph, resulting in significant performance improvements. Furthermore, Hamilton et al. \cite{hamiltoninductive2017} were the first to identify the latent nodes within a graph, leading to the proposal of the inductive spatial graph neural network known as GraphSAGE. As research progresses, scholars have observed that the modeling of real-life scenes using graphs typically undergoes dynamic evolution, extending beyond static data \cite{yuan2024environment, gao2024etc}. Based on distinct data forms (discrete and continuous), current research can be categorized into two groups \cite{Gravina24Deep}: discrete dynamic graph neural networks (DTDG), exemplified by DyTed \cite{zhang2023dyted} and SILD \cite{zhang2024spectral}, and continuous dynamic graph neural networks (CTDG), exemplified by Ada-DyGNN \cite{li2024robust} and SimpleDyG \cite{wu2024feasibility}. 

However, these approaches are limited to extracting information solely from dynamic graphs, lacking the capability to maintain memory and selective forgetting in the face of new data changes. BGML has made some breakthroughs in adapting to the inflow and withdrawal of graph data.

\subsection{Lifelong Learning and Graph Lifelong Learning}
Lifelong learning entails the continuous adaptation to new knowledge while refining existing knowledge over time \cite{ye2022Lifelong, Yang2023Cross}. In conventional domains, mainstream lifelong learning models can be broadly categorized into three main approaches. Firstly, architectural approaches often incorporate unit, extended, or compressed architectures to construct appropriate networks. For instance, ProgNN proposed by Rusu et al. \cite{rusu2016progressive} and DEN proposed by Yoon et al. \cite{yoon2017lifelong}. Secondly, regularization methods employ specialized loss terms to preserve the stability of previous parameters. Examples include EWC proposed by Kirkpatrick et al. \cite{kirkpatrick2017overcoming} and LWF developed by Li et al. \cite{li2017learning} Thirdly, rehearsal methods depend on a retraining process using samples from prior tasks. For instance, ICARL proposed by Rebuffi et al. \cite{rebuffi2017icarl}.

Recently, these strategies have been progressively extended into the domain of graph learning \cite{febrinanto2023graph}. Regarding architectural design, Wang et al. \cite{wang2022lifelong} devised the FGNs method to translate the graph structure into a format compatible with conventional data learning structures. To enhance the encoding of attribute information and the topological structure of target nodes, Zhang et al.'s \cite{zhang2022hierarchical} HPNs extract various levels of knowledge abstraction represented as atoms. TACO proposed by Han et al. \cite{han2024a} uses a topology-aware graph coarsening module to maintain the original graph topology information. Regarding the rehearsal scheme, Zhou et al. \cite{zhou2021overcoming} store past knowledge in an experience buffer and conduct lifelong learning on graph data through experience replay. Su et al. \cite{su2023towards} proposed SEA-ER by continuously using structural alignment to effectively combat catastrophic forgetting. In terms of regularization, Kou et al. \cite{kou2020disentangle} introduced decoupling-based continuous graph representation learning (DICGRL). Tan et al. \cite{tan2022graph} introduced GPIL with transferable meta-knowledge for initialization learning by introducing a pseudo-incremental approach. Furthermore, this field has given rise to several hybrid methods. ContinualGNN, proposed by Wang et al. \cite{wang2020streaming}, combines two approaches to alleviate catastrophic forgetting and preserve the existing learning mechanism. Feng et al. \cite{fengtowards} proposed a triple replay architecture (OTGNet) to solve the problem of heterogeneous propagation. The SEM proposed by Zhang et al. \cite{zhang2024ricci} combines the advantages of replay and architecture and performs sparse sampling through Ricci curvature.

Existing GLL methods focus on addressing catastrophic forgetting, aiming to retain as much old experience as possible while adapting to new knowledge. However, with limited resources, continuous information accumulation leads to performance loss, and these methods struggle when specific knowledge must be forgotten. In contrast, BGML focuses on enabling the model to selectively remember important information while forgetting irrelevant information. This selective memory strategy allows the model to efficiently discard outdated information while conserving memory space, thereby enhancing its ability to continue learning.

\subsection{Machine Unlearning and Graph Unlearning}
To ensure data privacy and model safety, the machine must learn to forget \cite{said2023survey, wu2023gif}. Cao et al. \cite{cao2015towards} were pioneers in introducing the concept of unlearning. Subsequently, researchers have proposed a series of new methods in two primary directions: exact unlearning and approximate unlearning \cite{liu2024model, kurmanji2024towards}. SISA (Sharded, Isolated, Sliced, and Aggregated), proposed by Bourtoule et al. \cite{Bourtoule2021Unlearning}, stands as one of the most representative works in this regard. The fundamental concept involves randomly dividing the training set into several disjoint shards and training the corresponding shard models independently. Upon receiving the forget request, the model only needs to retrain the respective shard model. Chatterjee et al. \cite{chatterjee2024unified} discussed the possibility of unifying continual learning with machine unlearning.

Recently, Chen et al. \cite{chen2022graph} extended the concepts of SISA to the graph domain, presenting GraphEraser as the inaugural graph unlearning model. Almost at the same time, Chien et al. \cite{chien2022certified} used the convex function characteristics for regularization and proposed certified graph unlearning. Building on this, Dong et al. \cite{dong2024idea} (IDEA) and Yi et al. \cite{yis2025calable} (ScaleGUN) each focus on improving flexibility and scalability, respectively. Wu et al. \cite{wu2023certified} constructed the edge unlearning framework CEU. Differently, GNNDelete proposed by Cheng et al. \cite{chenggnndelete} uses a model-independent hierarchical operator to optimize the deletion edge consistency and neighborhood influence for unlearning. Li et al. \cite{li2024towards} introduced MEGU to chieve a balance between forgetting performance and framework generalization. 

Current research on GUL remains primarily focused on static graphs, with limited exploration of how to adapt to environmental evolution during ``unlearning". BGML aims to apply this mechanism to dynamically evolving graphs, boldly addressing the memory issue in GML from the perspective of continuous increments and parallel forgetting.

\section{Problem  Formulation}\label{chap:3}
In GNNs, a static regular graph is typically depicted as $\mathcal{G}=\left ( \mathcal{V},\mathcal{E} \right )$, where $\mathcal{V}$ is the set of nodes and $\mathcal{E}$ is the set of edges in the graph. $N=\left |\mathcal{V} \right |$ signifies the count of nodes. The feature matrix of the graph is represented by $\mathbf{X} \in \mathbb{R} ^{N\times F \times C}$, where $F$ represents the feature dimension of each node. $C$ is the number of channels for each feature. Specifically, for nodes $v_{i}$ and $v_{j}$, their edge relationship can be denoted as $e_{i,j}$, while their respective characteristics are expressed as $\mathbf{X}_{i} \in \mathbb{R} ^{F \times C}$ and $\mathbf{X}_{j}  \in \mathbb{R} ^{F \times C}$. The adjacency matrix $\mathbf{A} \in \left \{ 0,1  \right \} ^{N\times N}$ encompasses all edge relationships within the graph.

The main target of GML is the dynamic evolution graph based on discrete snapshots, i.e., $\mathcal{G}=(\mathcal{G}_{0}, \mathcal{G}_{1}, \mathcal{G}_{2}, ..., \mathcal{G}_{t})$. The graph evolution process of each snapshot mainly includes two aspects. First, new data entering the graph network can be represented as $\mathcal{G}_{t}^{'} =  \mathcal{G}_{t-1} + \bigtriangleup \mathcal{G}_{t}^{'} = \mathcal{G}_{t-1}\cup \left \{ v_{\mathrm{new} }, e_{\mathrm{new}} \mid v_{\mathrm{new}}\in \mathcal{V}_{\mathrm{add}}   \right \}$, where $\mathcal{N}(v_{\mathrm{new}})$  represents the neighbors of $v_{\mathrm{new}}$ and $\mathcal{V}_{\mathrm{add}}$ represents the set of new nodes. Second, when the graph network receives forgetting requests (FR), the clearing operation can be expressed as $\mathcal{G}_{t}^{''} =  \mathcal{G}_{t-1} + \bigtriangleup \mathcal{G}_{t}^{''} = \mathcal{G}_{t-1}\setminus  \left \{ v_{\mathrm{forget} }, e_{\mathrm{forget}} \mid v_{\mathrm{forget}}\in \mathcal{V}  _{\mathrm{delete}}   \right \}  
$, where $\mathcal{V}_{\mathrm{delete}}$ is used to represent the set of nodes that need to be forgotten. In general, the change of the graph at time $t$ can be expressed as $\mathcal{G}_{t}= \mathcal{G}_{t-1}+ \bigtriangleup \mathcal{G}_{t}$, where $\bigtriangleup \mathcal{G}_{t}= \bigtriangleup \mathcal{G}_{t}^{'}+ \bigtriangleup \mathcal{G}_{t}^{''}$. In addition, task context changes over time throughout the process of graph memory learning. The task set can be represented by $\mathcal{T}=(\mathrm{T}_{0}, \mathrm{T}_{1}, \mathrm{T}_{2}, ... , \mathrm{T}_{t})$. The corresponding time is $t=(t_{0}, t_{1}, t_{2}, ... , t_{t})$. Notely, given the complexity of GML, this paper specifically focuses on the problem using nodes as an example to maintain clarity and depth. The scenario where only edges change during graph evolution has not been discussed in this work.

\begin{definition}\label{definition}
The objective of graph memory learning is to construct a model $H(\Theta)$ that continuously learns from graph data $\mathcal{G}=(\mathcal{G}_{0}, \mathcal{G}_{1}, \mathcal{G}_{2}, ... , \mathcal{G}_{t})$ by minimizing the model loss to achieve optimal performance on task $\mathrm{T}_{t}$ at time $t$. Additionally, the model should satisfy two conditions:

(i) It should possess the ability to alleviate the conflict between remembering and forgetting and overcome catastrophic forgetting, meaning it aims to maintain the performance of the model on tasks $\mathrm{T}_{1}$ through $\mathrm{T}_{t-1}$ as much as possible.

(ii) Nodes required to be forgotten at time $t_{t-1}$ should no longer have any relationship with the model at time $t_{t}$.
\end{definition}

\section{Graph Memory Learning Framework}\label{chap:4}
\begin{figure*}[!t]
	\centering
	\includegraphics[width=\textwidth]{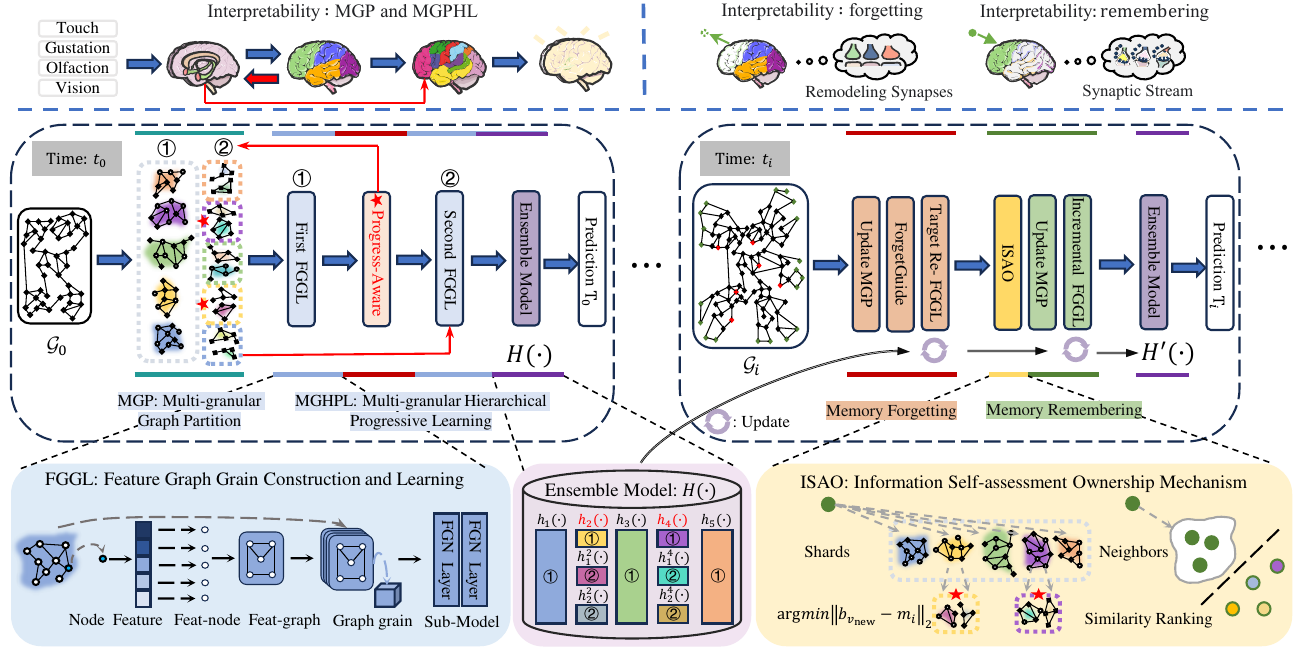}
	\caption{The overview of the BGML Framework. Similar to modular and hierarchical structures of the cerebral cortex, BGML perceives the muti-levels shards of $\mathcal{G}_{0}$ in a progressive manner. The results of the ensemble model $H(\cdot)$ show the graph learning ability of BGML. Subsequently, BGML imitates brain network dynamics and adapts to the dynamic evolution of the graph by dynamically updating the components in $H(\cdot)$. The blue dotted lines above correspond to brain network dynamics and function-structure coupling, reflecting the source of design inspiration at these two moments. The black dotted lines highlight the technical details of each important module in BGML. Notably, \textcircled{1}, \textcircled{2}, and the red mark at $t_{0}$-time concretely describe the concept of progressive learning.}
	\label{fig:overview}
\end{figure*}
\subsection{Morphological Dynamics of BGML in Temporal States}\label{sub:A}
According to Definition \ref{definition}, an algorithm framework adhering to the GML paradigm must complete the task set $\mathcal{T}=(\mathrm{T}_{0}, \mathrm{T}_{1}, \mathrm{T}_{2}, ... , \mathrm{T}_{i})$ at $t_{i}$-time. Moreover, the problem scenarios at $t_{0}$-time and $t_{i}$-time differ significantly. At $t_{0}$-time, the model needs to learn high-quality graph representations from the initial graph $\mathcal{G}_{0}$ to ensure accurate predictions for task $\mathrm{T}_{0}$. At $t_{i}$-time, the model must simultaneously handle incremental data requests (IR) and forgetting requests (FR) during the graph evolution process to complete $\mathrm{T}_{0}$ to $\mathrm{T}_{i}$.

As shown in Fig. \ref{fig:overview}, BGML is divided into two distinct forms at $t_{0}$-time and $t_{i}$-time. At $t_{0}$-time, BGML is based on the modular and hierarchical structure of the brain. Specifically, the initial graph $\mathcal{G}_{0}$ is divided into multi-level graph shards by multi-granular graph partition (MGP). These shards are then organized and progressively learned by the multi-granular hierarchical progressive learning mechanism (MGHPL) based on feature graph grain learning (FGGL). Finally, all the learned knowledge is integrated for prediction. At $t_{i}$-time, BGML builds a memory forgetting module and a memory remembering module, extending the mode at $t_{0}$-time to handle continuous data forgetting requests (FR) and incremental requests (IR). These modules are inspired by the brain's synaptic remodeling mechanism and 
synaptic stream mechanism. Furthermore, the information self-assessment ownership mechanism (ISAO) in the remembering module helps incremental knowledge form a credible graph structure.

It is crucial that the model at $t_{i}$-time is a continuation of the model at $t_{0}$-time along the timeline, rather than an independent design. The multi-granular and progressive understanding of local graph information by BGML at $t_{0}$-time is seamlessly inherited at $t_{i}$-time. This in-depth comprehension of changes in local and regional information significantly alleviates the conflict between remembering and forgetting instructions. At $t_{i}$-time, several designs, including ISAO, aim to protect existing knowledge and conserve computing resources during the model's continuous learning process. Emulating brain network dynamics, the forgetting and remembering modules at $t_{i}$-time continuously update the empirical knowledge in the integrated model from $t_{0}$-time, enabling continuous learning.

\subsection{MGP: Multi-granular Graph Partition}\label{sub:B}
Inspired by brain network principles such as the Sensory Pathways Theory and the Cortical Localization Theory (Remark \ref{remark1}), this paper proposes a multi-granularity graph partitioning approach. It mimics brain cognitive patterns to modularize and hierarchically decouple the graph knowledge.

\begin{remark}\label{remark1}
\textbf{The Cortical Localization Theory} posits that distinct regions of the cerebral cortex are specialized for processing different sensory, motor, and cognitive functions. The principles of Functional Localization and Regional Specialization suggest that specific areas of the cortex receive signals from particular sensory pathways. \textbf{The Sensory Pathways Theory} posits that sensory information is transmitted from external stimuli to the corresponding regions of the cerebral cortex for processing via specific neural pathways, such as the visual and auditory pathways. Furthermore, neuroimaging techniques (e.g. fMRI) have further confirmed that cortical division is precise and multi-level.
\end{remark}

\textbf{MGP Method.} In conclusion, the cerebral cortex's strategy of multi-region collaboration and hierarchical progression fosters the agent's profound comprehension of knowledge. This necessitates preprocessing before knowledge input to facilitate problem division and resolution. Inspired by this, the paper devised an MGP algorithm, As shown in the MGP module in Fig. \ref{fig:overview}. This algorithm aims to partition the static regular graph $\mathcal{G}_{0}$ at time $t_{0}$ into multiple granules, establishing a hierarchy of knowledge details. Subsequent graph evolution details will be simultaneously incorporated into the granule. This lays the groundwork for the subsequent in-depth understanding and hierarchical extraction of precise information in the dynamic evolution graph. Specifically, 
the MGP algorithm is built upon the subgraph partitioning methods BLPA and BEKM proposed by Chen et al. \cite{chen2022graph}. For $\mathcal{G}_{0}$, the two graph partitions of coarse and fine granularity can be expressed by Eq.\ref{eq: two partition}:
\begin{equation}\label{eq: two partition}
\begin{aligned}
&\left\{\begin{array}{l}
\mathrm{BLPA}(\mathcal{G}_{0})=\left \{ \mathcal{S}_{1},\mathcal{S}_{2},\dots ,\mathcal{S}_{k} \right \} \\  
\mathrm{BLPA}(\mathcal{S}_{i})=\left \{ \mathcal{P}_{1}^{i},\mathcal{P}_{2}^{i},\dots ,\mathcal{P}_{l}^{i} \right \},
\end{array}\right.  \\
&\left\{\begin{array}{l}
\mathrm{BEKM}(\mathcal{G}_{0})=\left \{ \mathcal{S}_{1},\mathcal{S}_{2},\dots ,\mathcal{S}_{k} \right \} \\  
\mathrm{BEKM}(\mathcal{S}_{i})=\left \{ \mathcal{P}_{1}^{i},\mathcal{P}_{2}^{i},\dots ,\mathcal{P}_{l}^{i} \right \}, 
\end{array}\right. 
\end{aligned}
\end{equation}
where $i=1,2,\dots,k$. $\mathcal{S}_{i}$ represents a first-level graph shard with coarse granularity. $\mathcal{P}_{j}^{i}$, on the other hand, denotes a second-level graph shard within $\mathcal{S}_{i}$, offering finer granularity, where $j=1,2,\dots,l$. $\mathrm{BLPA}(\cdot)$ and $\mathrm{BEKM}(\cdot)$ in Eq.\ref{eq: two partition} are abstract representations of iterative algorithms (i.e. Eq.\ref{eq: BLPA} and Eq.\ref{eq: BEKM}). Before iteration, they need to be provided with random initialized partition shards $\left \{ \mathcal{S}_{1},\mathcal{S}_{2},\dots ,\mathcal{S}_{k} \right \}^{0}$ and centroids $\left \{ \mathbf{m}_{1},\mathbf{m}_{2},\dots ,\mathbf{m}_{k} \right \}^{0}$, respectively. 

For $\mathrm{BLPA}(\cdot)$, the core iteration steps in which $\forall  v \in \mathcal{S}_{i}$ participates can be described as:
\begin{equation}\label{eq: BLPA}
\left\{\begin{array}{l}
\mathcal{S}_{\mathrm{src}} = \mathcal{S}_{i} \\  [6pt]
\mathcal{S}_{\mathrm{dst}} = \mathop{\arg\max}\limits_{\mathcal{S}_{i}, i=1,2,\dots,k}\left | \mathcal{N}_{\mathcal{S}_{i}}(v) \right | \\ [6pt]
\mathcal{S}_{\mathrm{dst}}\longleftarrow \mathcal{S}_{dst}\cup v \\ [6pt]
\mathcal{S}_{\mathrm{src}}\longleftarrow \mathcal{S}_{\mathrm{src}}\setminus  v,
\end{array}\right.
\end{equation}
where $\mathcal{S}_{\mathrm{src}}$ and $\mathcal{S}_{\mathrm{dst}}$, respectively, represent the shard where node $v$ is currently located and the shard to which it will belong. $\left | \mathcal{N}_{\mathcal{S}_{i}}(v) \right |$ is the number of neighbors of node $v$ in $\mathcal{S}_{i}$.

$\mathrm{BEKM}(\cdot)$ performs clustering by distance between node embedding and centroids. For $\forall  v$, the specific iteration steps can be expressed by Eq.\ref{eq: BEKM}:
\begin{equation}\label{eq: BEKM}
\left\{\begin{array}{l}
\mathcal{S}_{\mathrm{dst}} = \mathop{\arg\min}\limits_{\mathcal{S}_{i}, i=1,2,\dots,k}\left \| \mathbf{b}_{v} - \mathbf{m}_{i} \right \|_{2}  \\  [6pt]
\mathcal{S}_{\mathrm{dst}}\longleftarrow \mathcal{S}_{\mathrm{dst}}\cup v \\ [6pt]
\mathbf{m}_{\mathrm{dst}}\longleftarrow \frac{\sum_{v \in \mathcal{S}_{\mathrm{dst}}}\mathbf{b}_{v}}{\left | \mathcal{S}_{\mathrm{dst}}  \right | },
\end{array}\right.
\end{equation}
where $\mathbf{m}_{\mathrm{dst}}$ represents the centroid of the updated shard $\mathcal{S}_{\mathrm{dst}}$.

\subsection{MGHPL: multi-granular hierarchical progressive learning mechanism} \label{sub:C}
To address the frequent changes in local graph structures and mitigate the conflict between memory and forgetting, this paper, inspired by brain network principles such as the coordination and hierarchical processing of cortical regions, brain information fusion and decision-making (Remark \ref{remark2}), proposes a multi-granular hierarchical progressive learning mechanism based on the decoupling of graph knowledge.

\begin{remark}\label{remark2}
\textbf{The Coordination and Hierarchical Processing of Cortical Regions} refers to the coordinated functioning of different cortical areas during cognitive processes, progressing from primary sensory regions to secondary processing areas, and ultimately to higher-level cognitive integration and decision-making. This reflects the gradual deepening of the brain's cognitive processes. \textbf{The Brain Information Fusion and Decision-Making} refers to the extraction of information by modular and multi-level cortical regions, which is then fused and integrated to form higher-level cognitive features. It is crucial for accurate judgment and decision-making.
\end{remark}

\textbf{First FGGL.} Inspired by the above, this paper proposes the MGHPL mechanism based on feature graph grain learning (FGGL), as shown in Fig. \ref{fig:overview}. This mechanism takes the first-level graph shards and the second-level graph shards of the static graph at time $t_{0}$ given by the MGP algorithm as input. The first is the first-level FGGL. It is important that FGGL be related to subsequent continuous streaming graph learning as well as the response to node data IR and FR. This part of the idea refers to Wang et al.'s \cite{wang2022lifelong} implicit encoding of graph structure into the node feature graph, which breaks the shackles of node learning being restricted by the graph structure. Based on this, we also convert a single node into a feature graph and converting the graph into a collection of node feature graphs (i.e., \textbf{graph grains}). The adjacency matrix of the feature graph is constructed as follows:
\begin{equation}\label{eq: feature adjacency matrix}
\left\{\begin{array}{l}
\mathbf{A}_{c}^{\mathrm{feat}}(v_{i} )  = \mathrm{sgnroot}\left ( \frac{\sum_{v_{j}\in \mathcal{N}\left ( v_{i} \right )} \omega _{i,j}^{c} \mathbf{X}_{i}^{c} \left ( \mathbf{X}_{j}^{c} \right ) ^{\top }  }{\left | \mathcal{N}\left ( v_{i} \right )   \right | }  \right )    \\ 
\omega_{i,j}^{c} = \frac{\mathrm{exp}\left ( a_{i,j}^{c}  \right )}{\sum_{v_{r}\in \mathcal{N}\left ( v_{i} \right )}\mathrm{exp}\left ( a_{i,r}^{c}  \right )}   \\ [13pt]
a_{i,j}^{c} = \mathrm{LeakyReLU}\left ( \mathbf{w}_{i}^{\top }\mathbf{X}_{i}^{c} + \mathbf{w}_{j}^{\top }\mathbf{X}_{j}^{c} + q \right ) ,
\end{array}\right.
\end{equation}
where $\mathrm{sgnroot}(x) = \mathrm{sign}(x)\sqrt{\left | x \right | }$ and $\mathbf{X}_{i} \in \mathbb{R} ^{F \times C}$. For $\forall c \in \left \{ 1,2,\dots ,C \right \}$, $\mathbf{X}_{i}^{c} \in \mathbb{R}^{F}$ represents all features in the $c$-th channel of node $v_{i}$. $\omega_{i,j}^{c}$ represents the weight of the $c$-th channel edge $e_{i,j}$. $\mathbf{w}_{i},\mathbf{w}_{j} \in \mathbb{R}^{F}$, $q\in \mathbb{R}$ are learnable attention parameters. At this point, all first-level graph shards are converted into graph grains according to Eq.\ref{eq: feature adjacency matrix}. To avoid confusion, graph grains and graph shards share the same symbol: 
\begin{equation}\label{eq: grain}
\left \{ \mathcal{S}_{1},\mathcal{S}_{2},\dots ,\mathcal{S}_{k} \right \}_{\mathrm{shards}} \to \left \{ \mathcal{S}_{1},\mathcal{S}_{2},\dots ,\mathcal{S}_{k} \right \}_{\mathrm{grains}}.
\end{equation}

Then, each first-level graph grain is transferred as knowledge into the first-level sub-model $\left \{h_{1} (\cdot ), h_{2} (\cdot ), \dots, h_{k} (\cdot )\right \}$ of a different division of labor. These models are built to realize the function of cerebral cortex partitions. They form high-quality node embeddings $\left \{\mathbf{z}_{1}, \mathbf{z}_{2}, \dots, \mathbf{z}_{N}\right \}$ from the knowledge in graph grains for downstream tasks. Each sub-model adopts the same configuration and consists of two layers of feature graph networks \cite{wang2022lifelong}. For $\forall v_{i} \in \mathcal{S}_{j}$, The learning process of node embedding can be expressed by Eq.\ref{eq: first-level embedding}:
\begin{equation}\label{eq: first-level embedding}
\mathbf{z}_{i} = h_{j} (v_{i}) = \sigma \left ( \hat{\mathbf{A}}^{\mathrm{feat}} \sigma  \left (  \hat{\mathbf{A}}^{\mathrm{feat}} \mathbf{X}_{i} \mathbf{W} \right )\mathbf{W} \right ),
\end{equation}
where $\sigma(\cdot)$ is a non-linear activation function, $\mathbf{W}$ is a learnable parameter, and $\hat{\mathbf{A}}^{\mathrm{feat}}$ is the normalized feature graph adjacency matrix.

\textbf{Progress-aware Module.} According to the above, the first-level graph grains are constructed according to the coarse-grained first-level graph shards. However, when completing downstream tasks, some sub-models need to deeply understand the multi-level content and local details of the target. The sub-model trained according to the first-level graph grains obviously cannot fully meet this requirement. Therefore, this paper designs a progress-aware module based on the modular and hierarchical structure of brain cognition, which is used to screen first-level sub-models whose representation capabilities need further support. It is worth mentioning that the complete Validation dataset $\mathcal{G}_{Valid}$ is used to test the representation ability of each sub-model. The progress-aware module can be expressed by Eq.\ref{eq: progress-aware module}:
\begin{equation}\label{eq: progress-aware module}
\left\{\begin{array}{l}
\mathrm{score}_{i} = \mathrm{Predict} \left ( h_{i}\left ( \mathcal{G}_{Valid}   \right )   \right )  \\  [6pt]
\mathrm{score} = \left \{ \mathrm{score}_{1}, \mathrm{score}_{2}, \dots ,\mathrm{score}_{k} \right \} \\ [6pt]
\mathrm{\tau_{idx}} = \mathrm{Low_{rank}}\left ( \mathrm{score}, \tau \right ) ,
\end{array}\right.
\end{equation}
where $\mathrm{Predict}(\cdot)$ is a function that predicts task scores. $\mathrm{score}_{i}$ is the score whose output is the $i$-th first-level sub-model. $\mathrm{Low_{rank}}\left ( \mathrm{score}, \tau \right)$ is a retrieval function that can retrieve the sub-model index with the $\tau$-th score from the bottom.

The innovation of MGHPL lies in the gradual adaptation of the learning granularity, which mirrors the human cognitive process. Progress-aware module is the key technology for achieving this goal, which evaluates and provides real-time feedback on the cognitive ability of each sub-model, dynamically adjusting the integration framework to facilitate the incremental supplementation and integration of cognitive information at different granularities. This progressive, dynamically adjustable multi-granularity learning system enhances the model's ability to accurately delineate the boundary between retained and forgotten knowledge.

\textbf{Second FGGL.} The support method for the perceived sub-models is that the corresponding fine-grained second-level graph shards are constructed into second-level graph grains $\left \{ \mathcal{P}_{1}^{\tau_{idx}},\mathcal{P}_{2}^{\tau_{idx}},\dots ,\mathcal{P}_{l}^{\tau_{idx}} \right \}_{\mathrm{shards} } \to \left \{ \mathcal{P}_{1}^{\tau_{idx}},\mathcal{P}_{2}^{\tau_{idx}},\dots ,\mathcal{P}_{l}^{\tau_{idx}} \right \}_{\mathrm{grains}}$ to establish a more refined learning. For $\forall v_{i}\in \mathcal{P}_{j}^{\tau_{idx}}$ , the learning process of node
embedding can be expressed as:
\begin{equation}\label{eq: support}
\mathbf{z}_{i}^{'}  = h_{j}^{\tau_{idx}}\left ( v_{i}  \right ),
\end{equation}
where $h_{j}^{\tau_{idx}}$ and $\mathbf{z}_{i}^{'}$, respectively, represent the second-level sub-model corresponding to graph grain $\mathcal{P}_{j}^{\tau_{idx}}$ and its generated second-level embedding.

\textbf{Ensemble Model.} Ultimately, we unified the knowledge and experience learned from MGHPL in an ensemble way. A reasonable aggregation strategy can help the ensemble model make the correct decisions and judgments. There are three commonly used aggregation strategies: MajAggr based on majority voting, MeanAggr based on posterior averaging, and LBAggr based on learning \cite{chen2022graph}. These aggregation strategies are all suitable for the framework of this paper. The only thing that needs attention is the support strength of the second-level sub-model for the first-level sub-model that needs help. The ensemble strategy of MGHPL can be abstractly summarized:
\begin{equation}\label{eq: decision}
\left\{\begin{array}{l}
\chi _{\mathrm{final} } =  \mathrm{Aggr} \left \{\alpha_{1}, \alpha_{2},\dots ,\alpha_{k} \right \}   \\ [6pt]
\alpha_{\tau _{\mathrm{idx}}} =  \mathrm{Aggr} \left \{\alpha_{\tau _{\mathrm{idx}}}, \beta_{1}^{\tau _{\mathrm{idx}}}, \beta_{2}^{\tau _{\mathrm{idx}}},\dots ,\beta_{l}^{\tau _{\mathrm{idx}}} \right \} ,
\end{array}\right.
\end{equation}
where $\mathrm{Aggr}(\cdot)$ represents any of the aggregation strategies mentioned above. $\chi _{\mathrm{final} }$ denotes the final decision result. $\left \{\alpha_{1}, \alpha_{2},\dots ,\alpha_{k} \right \}$ contains decision information from all first-level sub-models, while $\left \{\beta_{1}^{\tau _{\mathrm{idx}}}, \beta_{2}^{\tau _{\mathrm{idx}}},\dots ,\beta_{l}^{\tau _{\mathrm{idx}}} \right \}$ encompasses decision information from all second-level sub-models associated with the first-level sub-model $h_{\tau _{\mathrm{idx}}}$ that needs support.

\subsection{Memory Forgetting Module}\label{sub:D}
The principles of autonomous synaptic selection, synaptic remodeling, and neural resource allocation in the brain's information forgetting (Remark \ref{remark3}) have inspired our approach to handling continuous forgetting request inputs. The memory forgetting module, driven by the forgetting-guided module, implements an efficient and localized pinpoint clearing model.

\begin{remark}\label{remark3}
\textbf{Autonomous Synaptic Selection} refers to the brain's ability to automatically identify and adjust the corresponding synaptic connections when confronted with erroneous information. \textbf{Synaptic Remodeling} involves the reorganization or restructuring of neural synapses in relevant regions during the forgetting process. This adjustment enables accurate forgetting. \textbf{Neural Resource Allocation} refers to the redistribution of neural resources during forgetting or synaptic remodeling, allowing the brain to focus on processing more meaningful information or skill acquisition, thereby enhancing learning efficiency.
\end{remark}

\textbf{Update MGP During Forgetting.} According to Definition \ref{definition} of GML, the key to dynamic graph evolution lies in continuous data increment and deletion. From time $t_{i-1}$ to $t_{i}$, the system faces two challenges: incremental new information requests and data forgetting requests. As shown in Fig. \ref{fig:overview}, to ensure that the learning of new information is not affected by the information that needs to be forgotten, the system should prioritize FR at each moment. Specifically, when the system receives FR, the update method of the corresponding graph shards can be expressed as:
\begin{equation}\label{eq: graph partition update}
\left\{\begin{array}{l}
\mathcal{S}_{i} \gets \mathcal{S}_{i}\setminus  \left \{ v_{\mathrm{forget} }, e_{\mathrm{forget}}  \right \}      \\ 
\mathcal{P}_{j}^{i} \gets \mathcal{P}_{j}^{i}\setminus  \left \{ v_{\mathrm{forget} }, e_{\mathrm{forget}}   \right \}  ,
\end{array}\right.
\end{equation}
where $v_{\mathrm{forget}}\in \mathcal{V}  _{\mathrm{delete}}$, $i=1,2,\dots,k$ and $j=1,2,\dots,l$.

\textbf{ForgetGuide Module.} All memories of the system at time $t_{i-1}$ are stored in the neuron parameters of each sub-model. When performing a forget operation, we only need to selectively lock the sub-models associated with the FR. The remaining irrelevant models maintain the state at time $t_{i-1}$. Therefore, forgetting guidance is very necessary. What deserves special attention is the compatibility between forgetting guidance and hierarchical, progressive decision-making strategies. In other words, it is crucial to determine whether the first-level sub-model, previously used to train the data that needs to be forgotten, is part of the sub-model perceived by the progress-aware module. If so, then the corresponding second-level sub-models also need to receive focused attention. The guidance operation can be simply expressed as:
\begin{equation}\label{eq: forgetting guidance}
\mathrm{Models} \left (v_{\mathrm{forget}}  \right ) = \mathrm{Guidance} \left ( v_{\mathrm{forget} }  \right ) ,
\end{equation}
where $ \mathrm{Guidance} \left ( \cdot \right )$ is the forgetting guidance function, which is used to guide all sub-models (first-level or second-level sub-models) related to the forgotten data.

\textbf{Re-FGGL.} Based on the updated graph shards of Eq.\ref{eq: graph partition update} and Eq.\ref{eq: feature adjacency matrix}, the graph grains have been updated in preparation for further learning. The irrelevant graph grains remain unaffected. For specific forgetting measures, this paper employs the scratch method. It solely involves pertinent sub-models, leaving the rest unaffected. Once the relevant sub-models have undergone retraining, they can seamlessly integrate into the decision-making process, serving as fresh references, thereby enhancing convenience.

\subsection{Memory Remembering Module}\label{sub:E}
The process of integrating new information with existing long-term memory involves not only the storage of information but also its reinforcement and consolidation. To address the conflicts arising from the low credibility of new knowledge structures, this paper, inspired by principles in brain dynamics such as cross-temporal memory comparison and integration, selective neural connection construction, and experience association (Remark \ref{remark4}), proposes the construction of a memory remembering module centered around ISAO. This module organizes reliable graph structure and buffers the conflicts caused by memory and forgetting.

\begin{remark}\label{remark4}
\textbf{Cross-temporal Memory Comparison and Integration} refers to the process of comparing current information with historical memories, integrating it based on the assessment of its variability or consistency. This process involves functional coordination between the prefrontal cortex and hippocampus. \textbf{Selective Neural Connection Formation} refers to the brain's ability to selectively establish new neural connections to avoid conflicts between new information and existing knowledge, facilitating the incorporation of new information into the memory system. \textbf{Experience Association} is the process through which the brain links and integrates new experiences with past ones, forming memories. This process aids in transforming new skills and information into long-term memory, enhancing their application and recognition.
\end{remark}

\textbf{ISAO Mechanism and Update MGP.} When new information enters the graph as a node, it meets neighbors and generates edges. In real-world scenarios, the introduction of incremental information may be relatively incidental. Arbitrarily assigning new nodes to a particular graph shard is unreasonable. This could result in a weak relationship between the new node and the entire graph shard or a lesser association with neighbors, thereby impacting the model's ability to learn and mine new knowledge. Therefore, faced with the influx of incremental information at the same time, this paper tends towards allowing new information to autonomously select the graph shard to which it belongs, as well as the most relevant neighbors. Specifically, we compute the distance between the new node and all centroids of graph shards and select the nearest first-level graph shard to incorporate the new node. This process can be represented by Eq.\ref{eq: centroids distance}:
\begin{equation}\label{eq: centroids distance}
\left\{\begin{array}{l}
d_{i} = \left \| \mathbf{b}_{v_{\mathrm{new}}} - \mathbf{m}_{i} \right \|_{2} \\ [8pt]
\mathcal{S}_{\mathrm{belong}} = \mathop{\arg\min}\limits_{i=1,2,\dots,k} \ d_{i},
\end{array}\right.
\end{equation}
where $\mathbf{b}_{v_{\mathrm{new}}}$ and $\mathbf{m}_{i}$ respectively denote the embedding of the new node and the centroid vector of the graph shard $\mathcal{S}_{i}$. $d_{i}$ represents the distance between them, and $\mathcal{S}_{\mathrm{belong}}$ denotes the shard to which the new node ultimately belongs. If the corresponding sub-model of the calculated first-level graph shard belongs to a sub-model perceived by the progress-aware module, it is necessary to reconfirm to which second-level graph shard the new node belongs.

Before commencing the cognitive processing of new data, selecting neighbors for the new node based on its ultimate affiliation is a crucial step. As depicted in Eq.\ref{eq: neighbors}, we calculate the similarity between the new node and all nodes within the shard (including other new nodes) and select the top-$\mu $ nodes with the highest similarity as the neighbors of the new node.
\begin{equation}\label{eq: neighbors}
\left\{\begin{array}{l}
 \eta = \psi  \left ( \mathbf{X}_{v_{new}},  \mathbf{X}_{i}  \right )  \\ [6pt]
\rho_{\mathrm{idx} } = \mathrm{Top} _{\mathrm{rank} }\left (\frac{1}{1+\eta},\mu   \right )  ,
\end{array}\right.
\end{equation}
where $\mathbf{X}_{v_{new}}$ and $\mathbf{X}_{i}$ respectively represent the features of the new node and other nodes within the shard. $\psi (\cdot)$ represents the Euclidean distance function based on the $L_{2}$-norm, used for computing the pairwise distance $\eta$ between nodes. $\frac{1}{1+\eta}$ inferred from distance calculation can characterize the similarity between nodes. $\mathrm{Top} _{\mathrm{rank} }(\cdot)$ function can extract the indices of the top-$\mu $ nodes with the highest similarity. $v_{\rho_{\mathrm{idx} }}$ denotes the neighbors of the new node. Additionally, it is worth noting that although incremental learning does not require a comprehensive update of the graph, the information regarding the new nodes and their edge relationships still needs to be updated within the graph shards. This is done to ensure that GML is not interrupted. Similar to the memory forgetting module, the graph shards in the MGP continue to be updated according to the ISAO mechanism.

ISAO simulates brain network dynamics to adaptively attribute and update knowledge, improving information reliability and consistency. Its key innovation is that new nodes select graph grains based on their features and adaptively adjust neighbor relationships, using the most relevant historical experiences to promote rapid information fusion and updating. This mechanism mimics how the human brain integrates new and old knowledge. The approach ensures the credibility of local graphs and neighboring nodes, while also adaptively aligning memory and unlearning requests with the corresponding sub-models of MGHPL, effectively mitigating the conflict between memory and forgetting.

\textbf{Incremental FGGL.} The determination of new node ownership provides corresponding sub-model index information. Furthermore, the construction of feature graphs and graph grains liberates nodes from the graph structure constraints, offering the possibility for independent incremental memory for new nodes. This implies that for incremental learning, we only need to focus on the new nodes, their neighboring information, and the relevant sub-models. The other components of the system remain inactive. The paired combination of new nodes and their neighboring information forms new samples, which are randomly streamed to the respective models. In other words, incremental learning is conducted on the corresponding sub-models based on Eq.\ref{eq: first-level embedding}. In the incremental task, each node is presented to the model only once, without reusing the same sample to update model parameters.

\subsection{The Algorithm Framework of BGML}\label{Sub: F}
\begin{algorithm}[h]
\caption{The algorithm for BGML.}\label{alg:alg1}
\renewcommand{\algorithmicrequire}{\textbf{Input:}}
\renewcommand{\algorithmicensure}{\textbf{Output:}}
\begin{algorithmic}[1]
        \REQUIRE Dynamic graph $\mathcal{G}=(\mathcal{G}_{0}, \mathcal{G}_{1}, ..., \mathcal{G}_{t})$, timestamps $t=(t_{0}, t_{1}, ... , t_{t})$, task set $\mathcal{T}=(\mathrm{T}_{0}, \mathrm{T}_{1}, ... , \mathrm{T}_{t})$, FR $\mathcal{V}_{\mathrm{delete}}$, IR $\mathcal{V}_{\mathrm{add}}$, number of first-level graph shards $k$, number of first-level graph shards $l$, number of first-level sub-models retrieved by the progress-aware module $\tau$, number of neighbors of new node $\mu $.  
        \ENSURE The final decision results of the tasks at all times $\chi _{\mathrm{final} }=(\chi_{\mathrm{final}}^{0}, \chi_{\mathrm{final}}^{1}, ... , \chi_{\mathrm{final}}^{t})$.    
        \FOR{$ \mathrm{timestamp} = t_{0} $ to  $t_{t}$}
            \IF {$ \mathrm{timestamp} = t_{0} $}
                \STATE Perform MGP (Eq.\ref{eq: two partition}) on $\mathcal{G}_{0}$  (Eq.\ref{eq: BLPA} or Eq.\ref{eq: BEKM}).
                \STATE First FGGL (Eq.\ref{eq: feature adjacency matrix}, Eq.\ref{eq: grain}, and Eq.\ref{eq: first-level embedding}).
                \STATE Perform progress-aware module (Eq.\ref{eq: progress-aware module}).
                \STATE Second FGGL (Eq.\ref{eq: support}).
                \STATE Ensemble model (Eq.\ref{eq: decision}),  predict task $\mathrm{T}_{0}$, and get the result $\chi_{\mathrm{final}}^{0}$.
            \ELSE
                \STATE Receive FR $\mathcal{V}_{\mathrm{delete}}$, update MGP(Eq.\ref{eq: graph partition update}).
                \STATE Perform ForgetGuide Module (Eq.\ref{eq: forgetting guidance}).
                \STATE Re-FGGL(Eq.\ref{eq: feature adjacency matrix}, Eq.\ref{eq: first-level embedding}, Eq.\ref{eq: support}).
                \STATE Receive IR $\mathcal{V}_{\mathrm{add}}$, determine ownership (Eq.\ref{eq: centroids distance}), select neighbors (Eq.\ref{eq: neighbors}), and update MGP.
                \STATE Incremental FGGL (Eq.\ref{eq: first-level embedding}, Eq.\ref{eq: support}).
                \STATE Get the final decision result $\chi_{\mathrm{final}}^{i}$ of task $\mathrm{T}_{i}$ (Eq.\ref{eq: decision}).
            \ENDIF
            \STATE Update timestamp $t_{i}\to t_{i+1}$.
        \ENDFOR
        \RETURN $\chi _{\mathrm{final} }=(\chi_{\mathrm{final}}^{0}, \chi_{\mathrm{final}}^{1}, ... , \chi_{\mathrm{final}}^{t})$
    \end{algorithmic}
\end{algorithm}

To enhance readers' comprehension of both graph memory learning and the content of this paper, this section systematically revisits the framework of the BGML algorithm, as shown in Alg.\ref{alg:alg1}. Within the BGML paradigm, we commence by processing and comprehending the static regular graph at time $t_{0}$, upon which all subsequent memory learning concerning dynamic evolutionary issues is predicated.

Firstly, at time $t_{0}$, the graph $\mathcal{G}_{0}$ is segmented into multiple shards using the balanced graph partitioning methods BLPA or BEKM. Secondly, multi-granular graph shards are fed into MGHPL based on progress-aware module for learning. Specifically, this paper transforms nodes into feature graphs by constructing a feature graph adjacency matrix, thereby converting graph shards into multi-granular graph grains. Subsequently, first-level graph grains are employed to train the first-level sub-model constructed by the feature graph networks. Moreover, these sub-models undergo evaluation and scoring. Based on the scores, the first-level sub-models requiring support can be progressively perceived, and corresponding second-level graph grain learning can be executed. Finally, the result $\chi_{\mathrm{final}}^{0}$ of task $\mathrm{T}_{0}$ can be derived by integrating sub-models following the hierarchical progressive strategy.

At time $t_{i}$ (not $t_{0}$), the system inherits all the information it possessed at time $t_{i-1}$. Upon receiving FR, the system removes the forgotten nodes from the graph shards. Relevant sub-models are then retrained to ensure complete clearing of associated memory. The second step involves receiving and learning IR. Firstly, the new nodes autonomously select which shard to belong to based on the distance between themselves and the centroid of each shard. Secondly, its neighbors are identified using the $L_{2}$ Euclidean distance between nodes. Subsequently, the graph shards are updated accordingly. Notably, new nodes and their neighbors constitute independent samples that are streamed into the system for incremental learning. Finally, the result $\chi_{\mathrm{final}}^{i}$ of task $\mathrm{T}_{i}$ is determined using the same decision-making strategy as at time $t_{0}$.

\section{Experiments and Analysis}\label{chap:5}
\subsection{Datasets and Baselines}
\textbf{Datasets.} Here we evaluate the advantages, functionality, effectiveness, and rationale of the BGML framework across nine node classification datasets. These datasets include widely-used benchmark datasets—Cora, CiteSeer, PubMed, Coauthor-CS, and Coauthor-Physics \cite{tu2024attribute, deng2024self}—as well as real-world datasets, namely Ogbn-arXiv \cite{wang2022lifelong}, Reddit \cite{hamiltoninductive2017}, Flickr \cite{wang2021mixup} and Yelp \cite{wang2021mixup}. Detailed information regarding these datasets is provided in Table \ref{Table_1}. The rest of the information about these datasets is presented as follows:
\begin{table}
\caption{Basic information of the datasets.}
\label{Table_1}
\centering
\resizebox{0.9\columnwidth}{!}{
\begin{tabular}{m{2cm}<{\centering}|m{1cm}<{\centering}m{1cm}<{\centering}m{1cm}<{\centering}m{1cm}<{\centering}m{1cm}<{\centering}m{1cm}<{\centering}m{1cm}<{\centering}}
    \toprule [1pt]
    Dataset     & Nodes & Edges & Features & Classes \\ [2pt] \midrule [1pt]
    Cora           & 2708  & 5429  & 1433     & 7       \\ [2pt] \midrule
    CiteSeer       & 3327  & 4732  & 3703     & 6       \\ [2pt] \midrule
    PubMed         & 19717 & 44338 & 500      & 3       \\ [2pt] \midrule
    Coauthor-CS    & 18333 & 81894 & 6805     & 15      \\ [2pt] \midrule
    Coauthor-Physics & 34493 & 247962 & 8415   & 5       \\ [2pt] \midrule [1pt]
    Ogbn-arXiv     & 169343 & 1166243 & 128     & 40      \\ [2pt] \midrule
    Reddit         & 232965 & 11606919 & 602     & 41      \\ [2pt] \midrule
    Flickr     &  89250 &  899,756 & 500     & 7      \\ [2pt] \midrule
    Yelp         & 716847 & 6977410 & 300     & 100      \\ [2pt] \bottomrule [1pt]
\end{tabular}}
\end{table}

\begin{itemize}
\item \textit{Citation networks}: Cora, CiteSeer, and PubMed \cite{tu2024attribute}. These datasets capture reference relationships among various documents, wherein each document is depicted as a node within a graph, and the connections between nodes represent citation links. Documents are represented as bags of words, hiding the characteristics of the category labels to which they belong.

\item \textit{Coauthor networks}: Coauthor-CS and Coauthor-Physics \cite{deng2024self}. These datasets primarily consist of co-authorship networks. Each author is represented as a node in the graph, and the connections between authors who co-authored a paper form the edges. Node characteristics are denoted by the keywords from the author's papers, while the labels signify the author's primary research field.

\item \textit{Real-world datasets}: Ogbn-arXiv \cite{wang2022lifelong}, Reddit \cite{hamiltoninductive2017}, Flickr \cite{wang2021mixup} and Yelp \cite{wang2021mixup}. Both datasets are sourced from real-world scenarios and serve as representative examples of large graph datasets. The Ogbn-arXiv dataset represents the citation network of all computer science-related papers on the arXiv platform. Notably, the Reddit dataset is a dynamic graph derived from the social news website Reddit, encompassing users from various communities and their interactions. In the Flickr network, nodes represent individual images. An edge is created between two images if they share common characteristics, such as the same geographic location or gallery. The Yelp dataset contains a recommendation network, where an edge signifies that the connected users are friends.
\end{itemize}

\textbf{Baselines.} Considering the novelty and complexity of graph memory learning, several factors were taken into account when choosing the baseline method for comparative experiments. For comparison, we selected two categories of methods. The first includes classic and widely-used methods in the current graph field, such as MLP, ChebNet \cite{defferrardconvolutional2016}, GCN \cite{TKipf2017}, GraphSAGE \cite{hamiltoninductive2017}, GAT \cite{velivckovicgraph2018}, APPNP \cite{Klicpera2019}, and Mixup \cite{wang2021mixup}. It is worth noting that when GCN and GAT are inconvenient to use with large-scale datasets, we opt to combine them with GraphSAINT \cite{zenggraphsaint} before making comparisons. The second category comprises methods related to graph memory learning, namely the graph lifelong learning leader methods: FGNs \cite{wang2022lifelong}, SEA-ER \cite{su2023towards} SEM \cite{zhang2024ricci}, OTGNet \cite{fengtowards}, TACO \cite{han2024a}, and the graph unlearning pioneer methods: GraphEraser \cite{chen2022graph}, SGC-Unlearn \cite{chien2022certified}, GraphDelete \cite{chenggnndelete}. 

The choice was made as GML is a novel subject that is only first discussed in this work, and no previous works have been found to serve as baselines for comparative experiments. The GML described in Definition \ref{definition} must, in essence, be capable of machine unlearning, continuous learning, and strong graph representation at all times. As a result, we chose the classic work in the aforementioned graph learning field that is most pertinent to these capability requirements for conducting comparative experiments to test model capabilities. We selected competitive methods from the domains of GUL and GLL for comparative experiments, based on the specific adaptability requirements of different tasks. Unfortunately, methods that are not suitable for GML problems could not be included in the comparison. Notely, the model-agnostic graph methods that appear in this paper, such as GraphEraser, are all based on GCN to form a specific model.

\subsection{Experiment Settings}
\textbf{Settings.} This subsection outlines the standardized setup used for all experiments in this paper. For graph partition, this paper uses two levels of granularity partitioning as an example. If necessary, more levels of granularity partitioning can be performed because MGP satisfies the binary tree partitioning criterion. Specifically, we adopted the settings from the BEKM and BLPA algorithms as described in the literature \cite{chen2022graph}. To determine the number of shards at each level, we used the number of cerebral cortex partitions as a reference and employed the grid search method for selection. The number of shards is correlated with the dataset size, and adjusting it appropriately can enhance the overall model performance. To prevent excessive focus on minor details that might affect the verification of the overall model function, we standardized the number of first-level shards ($k=5$) and second-level shards ($l=2$) in the experiments. For model training, we employed the SGD optimizer. The number of sub-models retrieved by the progress-aware module ($\tau=2$) and the number of independently selected neighbors for incremental information ($\mu=3$) were empirically determined through the grid search method in this study. The evaluation metric used in this study is the harmonic mean of precision and recall, known as the Micro F1-Score. Historically, this metric has been commonly employed to assess the multi-classification performance of GNN models.

\textbf{Datasets Division for Tasks.} Initially, we utilized a random division method to split all datasets into separate training and test sets in an 0.8:0.2 ratio. For \textbf{regular} graph learning and graph \textbf{unlearning} tasks, all training graphs were directly fed into the model for training. However, in scenarios involving \textbf{data-incremental} and \textbf{class-incremental}, as well as graph \textbf{memory} issues, the training set was divided into an initial training graph (at time $t_{0}$) and random incremental data input through streaming. The division ratio and the number of incremental data per time step can be adjusted flexibly based on task requirements (\ref{chap:5-H}). Similarly, the number of FR can be adjusted accordingly (\ref{chap:5-H}). Specifically, Fig. \ref{fig: real-world} presents the dynamic experiments on the applicability of BGML and various baselines to the GML problem across real-world datasets. The experiments include the large-scale graph dataset Ogbn-Arxiv, as well as three dynamic networks with natural temporal information (Reddit, Flickr, and Yelp). The partitioning of Ogbn-Arxiv follows the same setup as other commonly used node classification datasets. For the remaining three dynamic networks with timelines, we select the midpoint of the overall timeline as the $t_{0}$ moment for the GML problem. The subsequent incremental nodes and the dynamic changes in the graph correspond to the natural temporal evolution of the dataset, with forgotten nodes being randomly selected. The number of FR can be adjusted flexibly, for example, 50 or 100 nodes per round. For clarity, the line graph in Fig. \ref{fig: real-world} reports the performance at $t_{0}$ and at four equidistant time nodes ($t_{1}$ to $t_{4}$) on the remaining timeline.

\textbf{Implementation.} The experiments about BGML were conducted in a virtual environment utilizing Python 3.9.18 and PyTorch 1.11. The experiments were conducted on a server equipped with an NVIDIA GeForce RTX 4090 GPU. The server boasts 24GB of memory and runs on the Ubuntu 22.04.2 LTS operating system. All experiments about model performance were repeated 10 times. The mean and standard deviation of these experiments were calculated and reported.

\subsection{Evaluation of Learning and Memory Efficacy}
\begin{table*}
\centering
\caption{Regular graph learning (F1-score / \%). Bold fonts highlight the optimal performance on each dataset in this experiment, while suboptimal results are underlined.}
\label{Table_2}
\begin{tabular*}{\textwidth}{p{2cm}<{\centering}|p{2cm}<{\centering}|p{2.2cm}<{\centering}p{2.2cm}<{\centering}p{2.2cm}<{\centering}|p{2.2cm}<{\centering}p{2.2cm}<{\centering}} 
\cline{1-3}\cline{3-5}\cline{5-7}
\toprule [1pt]
Model/Dataset & Task Type                & Cora             & CiteSeer         & PubMed           & Coauthor-CS      & Coauthor-Phy      \\ [2pt] 
\midrule [1pt]
MLP           & \multirow{10}{*}{\makecell[c] { Regular \\ \\ (None)}} & 65.72±1.54       & 69.35±1.89       & 78.66±2.20       & 90.07±1.33       & 91.72±2.04        \\[2pt] 
ChebNet       &                          & 80.33±0.72       & 71.49±2.13       & 82.41±1.42       & 90.74±0.88       & 93.54±0.68        \\[2pt]
GCN           &                          & 83.27±0.64       & 73.35±0.83       & 86.04±0.58       & 93.25±0.72       & 94.70±0.44        \\[2pt]
GraphSAGE     &                          & 80.68±0.54       & 71.64±0.66       & 85.47±0.38       & 92.93±0.42       & 94.18±0.57        \\[2pt]
GAT           &                          & 83.69±0.43       & 73.68±0.48       & 86.20±0.26       & 93.86±0.33       & 95.52±0.21        \\[2pt] 
APPNP         &                          & 84.44±0.36       & 76.21±0.08       & 87.23±0.23       & 94.34±0.54       & 95.19±0.32        \\[2pt] 
FGNs          &                          & \underline{86.80±0.45}       & \underline{78.14±0.33}       & \underline{88.23±0.36}       & \underline{95.24±0.16 }      & \underline{96.14±0.25 }       \\[2pt]
GraphEraser   &                          & 83.85±0.36       & 74.62±0.47       & 86.33±0.34       & 92.74±0.22       & 93.93±0.17        \\[2pt] 
\textbf{BGML}          &                          & \textbf{88.89±0.14}       & \textbf{78.79±0.20 }      & \textbf{90.20±0.04}       & \textbf{96.72±0.11}       & \textbf{97.36±0.23}        \\ [2pt]
\bottomrule [1pt]
\end{tabular*}
\end{table*}
\begin{table*}
\centering
\caption{Graph memory learning (F1-score / \%). Bold fonts highlight the optimal performance on each dataset in this experiment, while suboptimal results are underlined. ``\textbackslash" denotes the experiment that cannot be executed given the constraints of the task.}
\label{Table_3}
\begin{tabular*}{\textwidth}{p{2cm}<{\centering}|p{2cm}<{\centering}|p{2.2cm}<{\centering}p{2.2cm}<{\centering}p{2.2cm}<{\centering}|p{2.2cm}<{\centering}p{2.2cm}<{\centering}} 
\cline{1-3}\cline{3-5}\cline{5-7}
\toprule [1pt]
Model/Dataset & Task Type                & Cora             & CiteSeer         & PubMed           & Coauthor-CS      & Coauthor-Phy      \\ [2pt] 
\midrule [1pt]
MLP           & \multirow{12}{*}{\makecell[c] { Memory \\ \\ (FR + Data-IR)}}  & 63.44±1.68       & 68.32±1.59       & 77.93±2.14       & 88.13±1.37       & 87.66±1.83        \\[2pt] 
ChebNet       &                          & 79.20±0.55       & 70.42±1.92       & 81.25±1.29       & 89.34±0.66       & 92.47±0.72        \\[2pt] 
GCN           &                          & 81.13±0.24       & 73.30±0.74       & 83.32±0.38       & 92.75±0.52       & 93.57±0.39        \\[2pt] 
GraphSAGE     &                          & 77.76±0.37       & 70.96±0.46       & 82.12±0.44       & 91.11±0.33       & 93.16±0.54        \\[2pt] 
GAT           &                          & 82.62±0.46       & 72.84±0.35       & 84.46±0.32       & 93.12±0.20       & 94.65±0.34        \\[2pt] 
APPNP         &                          & 84.45±0.22       & 72.55±0.28       & 85.34±0.27       & 93.44±0.15       & 94.33±0.23        \\[2pt] 
FGNs          &                          & 85.13±0.34       & 75.54±0.36       & 86.42±0.16       & 94.58±0.09       & 95.04±0.28        \\[2pt] 
GraphEraser   &                          & \textbackslash{} & \textbackslash{} & \textbackslash{} & \textbackslash{} & \textbackslash{}  \\[2pt] 
SEA-ER          &                          & 85.89±0.23       & 76.04±0.32       & 87.16±0.21       & 94.43±0.14       & \underline{95.98±0.25}        \\[2pt] 
SEM          &                          & \underline{86.12±0.11}       & \underline{77.13±0.18}       & \underline{87.24±0.35}       & \underline{94.88±0.14}       & 95.12±0.16        \\[2pt] 
\textbf{BGML}          &                          & \textbf{86.30±0.15}      & \textbf{79.28±0.22}       & \textbf{88.46±0.08}       & \textbf{96.03±0.18}       & \textbf{96.58±0.26}        \\[2pt] 
\bottomrule [1pt]
\end{tabular*}
\end{table*}
To verify that BGML has the powerful ability to learn high-quality graph representations at any time, Table \ref{Table_2} presents a summary of the experimental results for BGML and several benchmark comparison methods for regular static graph node classification tasks. Upon observation of the table, it becomes evident that BGML, as proposed in this paper, outperforms all benchmark methods across various datasets. This not only demonstrates the suitability of BGML for conventional node classification experiments but also highlights its significant advantages in mining intrinsic graph information and learning high-quality graph representations. This capability serves as a fundamental assurance for the continued outstanding performance of models in subsequent dynamic evolutionary graphs.

As discussed in Definition \ref{definition}, the graph memory learning problem outlined in Table \ref{Table_3} necessitates the model's capability to manage IR and FR concurrently. However, the graph methods in the baselines are incapable of addressing this challenge amid graph evolution. In our experiments, we endeavored to adapt them to the graph memory learning problem using training methods such as streaming and Scratch. Unfortunately, GraphEraser, which is unsuitable for streaming, could not participate in the comparison. The data in the table clearly demonstrates the significant advantages of BGML in addressing graph memory learning problems. Notably, BGML boasts learning and training methods specifically designed to address the challenges of graph memory learning. Considering its exceptional performance, BGML emerges as the first comprehensive solution to the graph memory learning problem in the graph field. It possesses the capability to continuously remember and forget specified information without necessitating retraining. Furthermore, the noteworthy benefits of BGML over other methods demonstrate that our designed MGHPL method comprehends the evolving local graph knowledge in a progressive and multi-granular manner, effectively alleviating the conflict between remembering and forgetting instructions.

\subsection{Evaluation of Unlearning Efficacy}
To ensure the adaptability of the model to diverse graph memory learning scenarios, such as instances with no IR at certain times or only forgetting instructions, it is deemed necessary to separately evaluate the model's ability to unlearn. Table \ref{Table_4} presents the performance of each graph model after processing an equal number of random FR. Clearly, the BGML method proposed in this article outperforms other baseline methods across all datasets. This demonstrates that BGML can effectively clear the model's memory when dealing solely with FR while also generating high-quality graph embeddings.

\begin{figure}
	\setlength{\abovecaptionskip}{-0.5cm}
	\setlength{\belowcaptionskip}{0cm}
	\begin{center}
		\subfloat[Unlearning efficiency]{
			\includegraphics[width=0.245\textwidth, height=3.8cm]{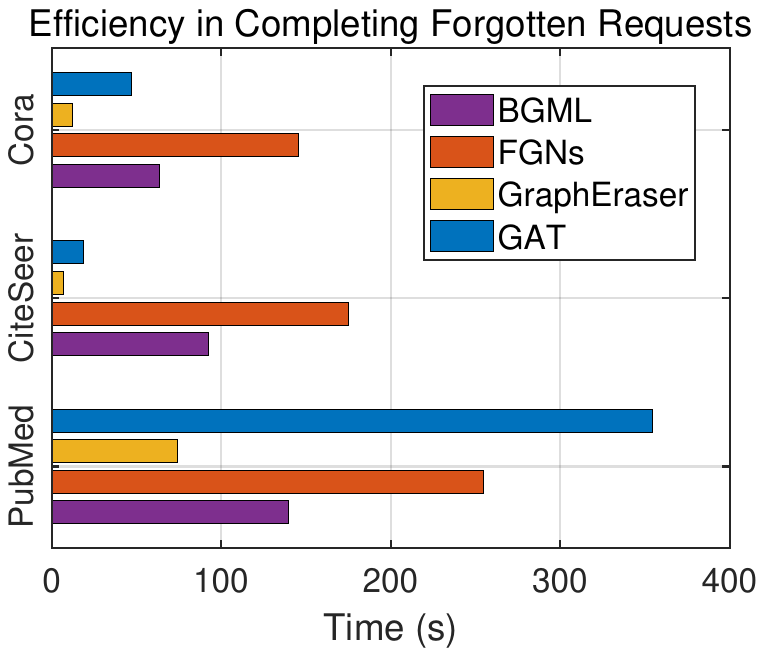}}
		\subfloat[Catastrophic forgetting]{
			\includegraphics[width=0.245\textwidth, height=3.8cm]{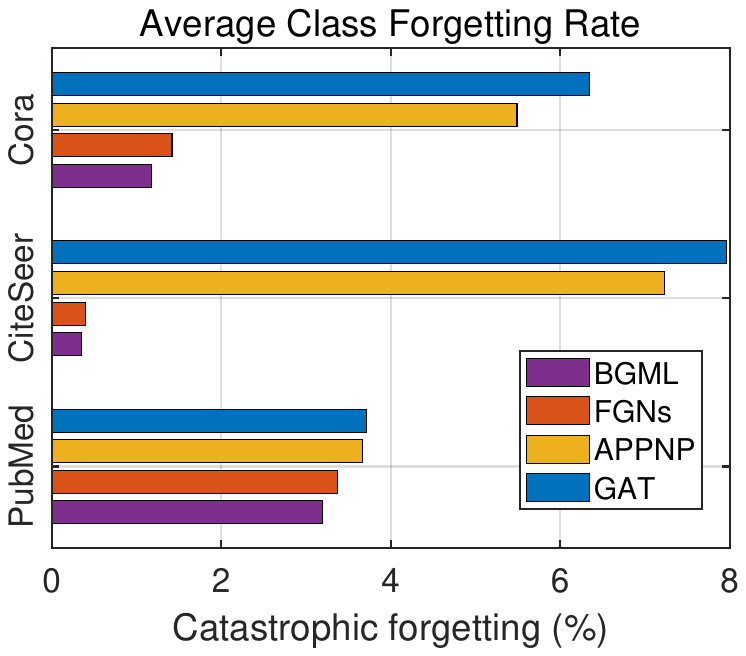}}
	\end{center}
        \vspace{0.35cm}
	\caption{Unlearning efficiency and catastrophic forgetting rate.}\label{fig: Forgetting requests and Catastrophic forgetting}
	
\end{figure}

\textbf{Unlearning efficiency.} Unlearning aims to conserve time and computing resources by avoiding model retraining when sensitive or harmful information is encountered. Therefore, it is important to initiate a discussion on forgetting efficiency. Fig. \ref{fig: Forgetting requests and Catastrophic forgetting}(a) shows that BGML's forgetting efficiency on the Cora, CiteSeer, and PubMed datasets is second only to GraphEraser, surpassing other models. The increased time cost compared to GraphEraser is attributed to the transformation of graph shards into feature graph-composed graph grains. Nevertheless, graph grain transformation aims to enhance persistent memory, a capability absent in GraphEraser. Essentially, BGML effectively balances unlearning efficiency and learning continuity.
\begin{table*}
\centering
\caption{Graph unlearning (F1-score / \%). Bold fonts highlight the optimal performance on each dataset in this experiment, while suboptimal results are underlined.}
\label{Table_4}
\begin{tabular*}{\textwidth}{p{2cm}<{\centering}|p{2cm}<{\centering}|p{2.2cm}<{\centering}p{2.2cm}<{\centering}p{2.2cm}<{\centering}|p{2.2cm}<{\centering}p{2.2cm}<{\centering}} 
\cline{1-3}\cline{3-5}\cline{5-7}
\toprule [1pt]
Model/Dataset & Task Type                & Cora             & CiteSeer         & PubMed           & Coauthor-CS      & Coauthor-Phy      \\ [2pt] 
\midrule [1pt]
MLP           & \multirow{12}{*}{\makecell[c] { Unlearning \\ \\ (FR)}} & 65.50±1.56  & 69.94±1.73 & 78.64±1.94 & 90.22±1.42  & 91.83±1.88        \\[2pt] 
ChebNet       &                          & 79.25±0.67 & 71.26±2.07 & 82.23±1.34 & 90.43±0.76  & 93.22±0.70        \\[2pt]
GCN           &                          & 82.86±0.50 & 72.69±0.85 & 85.27±0.44 & 93.02±0.65  & 94.38±0.47        \\[2pt]
GraphSAGE     &                          & 80.24±0.47 & 71.12±0.52 & 84.35±0.41 & 92.45±0.36  & 93.97±0.55        \\[2pt]
GAT           &                          & 83.82±0.48 & 73.50±0.52 & 85.64±0.23 & 93.72±0.34  & 95.35±0.17        \\[2pt] 
APPNP         &                          & 84.12±0.35 & 75.74±0.12 & 86.92±0.25 & 93.90±0.41  & 95.03±0.20        \\[2pt] 
FGNs          &                          & 86.24±0.40       & \underline{77.43±0.29}       & 87.67±0.38       & \underline{94.95±0.13}      & 95.88±0.16       \\[2pt]
GraphEraser   &                          & 81.49±0.53 & 74.10±0.45 & 85.94±0.26 & 92.53±0.14  & 93.56±0.24        \\[2pt] 
GraphDelete   &                          & \underline{86.37±0.62} & 77.28±0.14 & \underline{88.53±0.43} & 94.26±0.22  & \underline{96.10±0.54}        \\[2pt]
SGC-Unlearn   &                          & 84.34±0.41 & 76.31±0.35 & 86.57±0.15 & 92.55±0.23  & 93.26±0.47        \\[2pt]
\textbf{BGML}          &                          & \textbf{87.93±0.14}       & \textbf{77.58±0.18}      & \textbf{89.70±0.10}       & \textbf{96.50±0.23}       & \textbf{97.13±0.23}        \\ [2pt] 
\bottomrule [1pt]
\end{tabular*}
\end{table*}

\subsection{Evaluation of Incremental Learning Efficacy}
Similarly, in graph memory learning problems, instances arise where the system continues to process IR without receiving the FR instruction at a specific moment. This situation transforms the problem into incremental graph learning. To assess BGML's incremental learning capabilities, this paper conducted experiments on both standard data-incremental scenarios and class-incremental scenarios. The specific experimental results are presented in Table \ref{Table_5} for data-incremental scenarios and Table \ref{Table_6} for class-incremental scenarios. Upon closer examination of these tables, it is evident that BGML has demonstrated varying degrees of superiority in both scenarios and across all datasets. This demonstrates BGML's ability to maintain a high level of node embedding extraction during map incremental learning.

\textbf{Catastrophic forgetting problem.} Overcoming catastrophic forgetting in incremental learning has consistently been a significant challenge for researchers. In response to this challenge, we conducted experimental evaluations using class-incremental scenarios. Fig. \ref{fig: Forgetting requests and Catastrophic forgetting}(b) illustrates the average class forgetting rate of various models during continual learning on the Cora, CiteSeer, and PubMed datasets. In longitudinal comparison, BGML exhibits the lowest forgetting rate on the three datasets compared to the common baseline model. This demonstrates the greater reliability of our model's memory capability. This advantage stems from two main factors: firstly, BGML inherits the benefits of FGNs in continuous incremental learning; secondly, BGML's brain-like partition learning facilitates memory retention by isolating increments.

\subsection{Evaluation on Real-world Datasets}
\begin{figure}
	\setlength{\abovecaptionskip}{-0.5cm}
	\setlength{\belowcaptionskip}{0cm}
	\begin{center}
		\subfloat[Arxiv]{
			\includegraphics[width=0.245\textwidth, height=3.15cm]{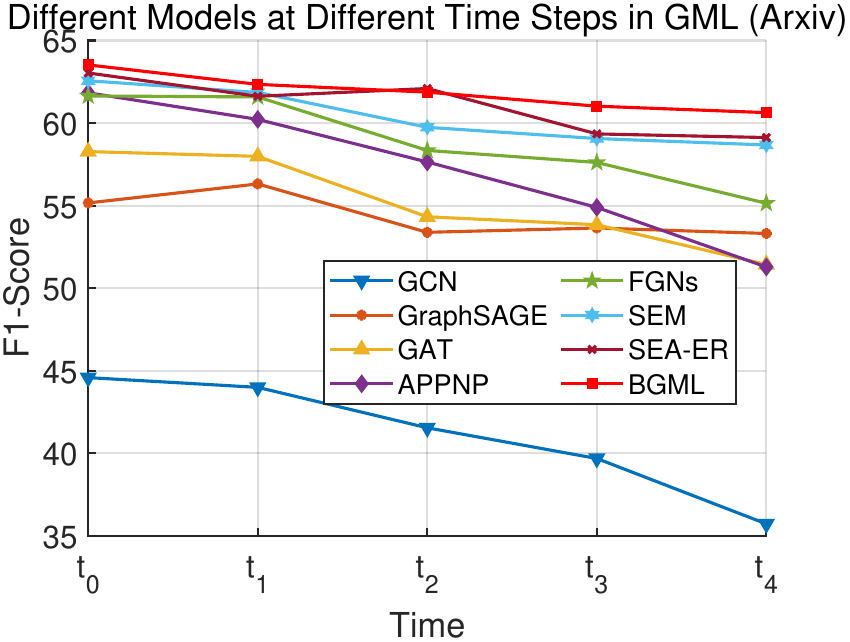}}
		\subfloat[Reddit]{
			\includegraphics[width=0.245\textwidth, height=3.15cm]{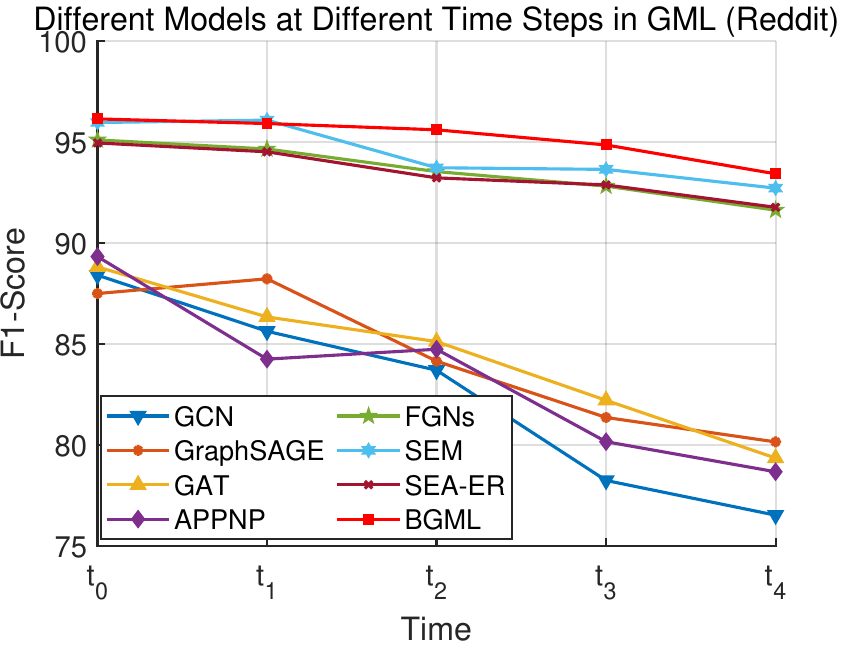}} \\
        \subfloat[Flickr]{
			\includegraphics[width=0.245\textwidth, height=3.15cm]{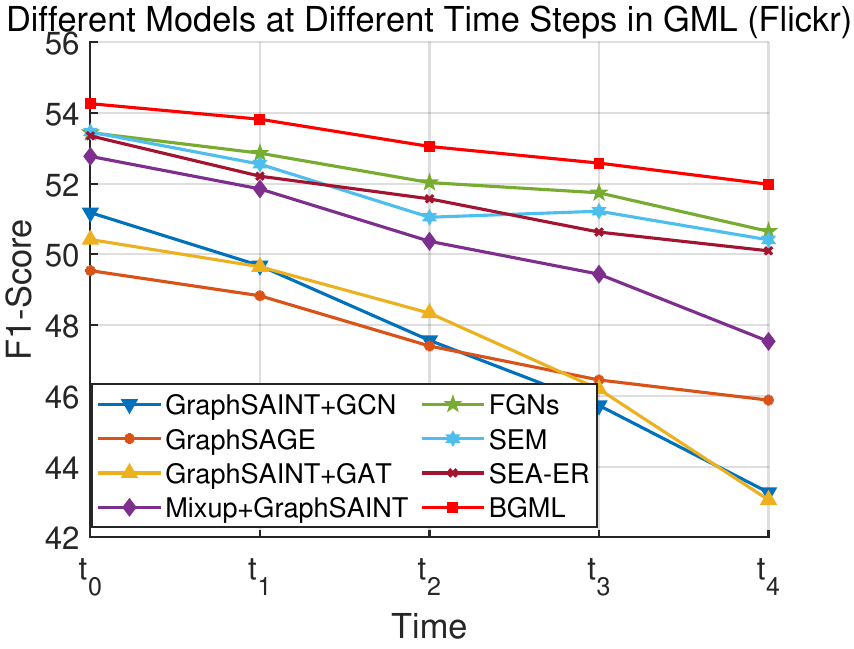}}
		\subfloat[Yelp]{
			\includegraphics[width=0.245\textwidth, height=3.15cm]{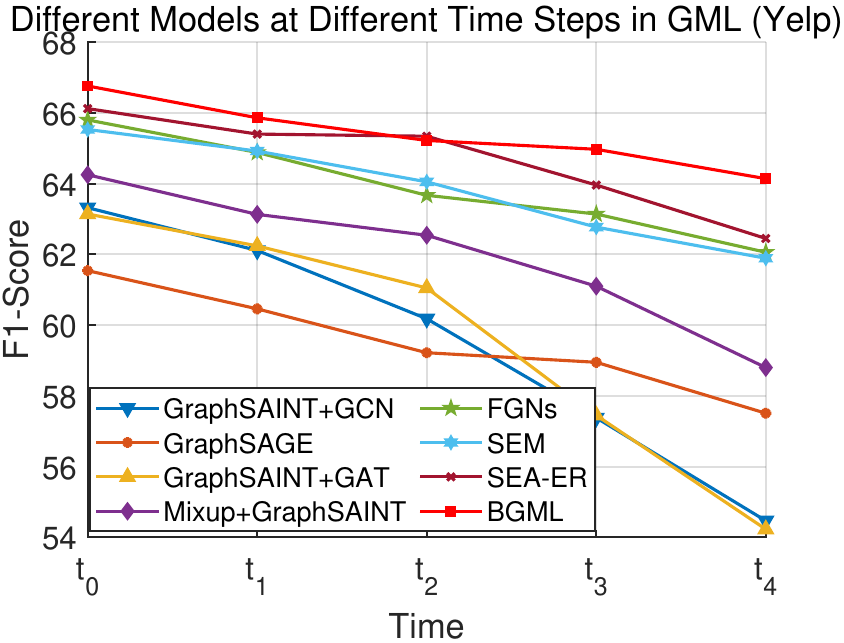}}
	\end{center}
        \vspace{0.35cm}
	\caption{Performance of different models at different time steps in GML.}\label{fig: real-world}
	
\end{figure}

In Fig.\ref{fig: real-world}, we present a dynamic visualization of the performance trends of BGML and several baseline models in solving GML problems across four real-world datasets using a timeline and line graph. The main goal of this experiment is to assess the advantages and unique capabilities of BGML in addressing the challenges of GML within large-scale, real-world dynamic networks that inherently include temporal information.

The experimental results provide compelling evidence that, across both datasets, BGML outperforms other models in terms of accuracy and stability throughout most periods of the GML task. This highlights BGML's strong capacity to handle GML problems continuously. More notably, when compared to traditional graph learning models and other baseline methods, BGML's line graph shows a more stable trend, thus demonstrating its robustness in dynamic, evolving environments. The findings further support the central argument of this paper: BGML effectively mitigates the conflict between "memory" and "forgetting." In contrast, other benchmark models suffer from significant performance fluctuations and degradation due to the frequent introduction and removal of new and old data. By incorporating MGHPL and ISAO, BGML efficiently processes incremental updates, maintains robust memory retention of historical information, while selectively forgetting certain details, even in the context of a constantly evolving graph.

\begin{table*}
\centering
\caption{Data-Incremental Learning (F1-score/\%). Bold fonts highlight the optimal performance on each dataset in this experiment, while suboptimal results are underlined. ``\textbackslash" denotes the experiment that cannot be executed given the constraints of the task.}
\label{Table_5}
\begin{tabular*}{\textwidth}{p{2cm}<{\centering}|p{2cm}<{\centering}|p{2.2cm}<{\centering}p{2.2cm}<{\centering}p{2.2cm}<{\centering}|p{2.2cm}<{\centering}p{2.2cm}<{\centering}} 
\cline{1-3}\cline{3-5}\cline{5-7}
\toprule [1pt]
Model/Dataset & Task Type                & Cora             & CiteSeer         & PubMed           & Coauthor-CS      & Coauthor-Phy      \\ [2pt] 
\midrule [1pt]
MLP           & \multirow{11}{*}{\makecell[c] {Data-Incremental \\  \\ (Data-IR)}} & 63.63±1.56       & 68.83±1.57       & 78.35±1.94       & 88.45±1.92       & 89.44±1.12        \\[2pt] 
ChebNet       &                          & 79.50±0.64       & 70.85±1.88       & 81.96±1.44       & 89.96±0.84       & 92.85±0.68        \\[2pt]
GCN           &                          & 81.44±0.48       & 73.24±0.74       & 83.76±0.36       & 92.87±0.43       & 93.94±0.47        \\[2pt]
GraphSAGE     &                          & 77.54±0.44       & 71.26±0.40       & 82.43±0.38       & 91.36±0.26       & 93.55±0.64        \\[2pt]
GAT           &                          & 83.05±0.52       & 72.97±0.35       & 84.88±0.26       & 93.33±0.23       & 94.92±0.31        \\[2pt] 
APPNP         &                          & 84.53±0.18       & 73.20±0.33       & 85.68±0.25       & 93.54±0.20       & 94.73±0.23        \\[2pt] 
FGNs          &                          & 85.44±0.32       & 75.97±0.34       & 87.24±0.26       & 94.73±0.12       & 95.36±0.16       \\[2pt]
SEA-ER          &                          & 86.88±0.26       & 76.34±0.19       & \underline{87.95±0.22}       & 95.01±0.13       & \underline{96.22±0.31}        \\[2pt] 
SEM          &                          & \underline{87.05±0.20}       & \underline{77.42±0.42}       & 87.56±0.28       & \underline{95.34±0.15}       & 95.97±0.25        \\[2pt] 
\textbf{BGML}          &                          & \textbf{87.14±0.14}       & \textbf{79.70±0.20}        & \textbf{88.88±0.04}       & \textbf{96.44±0.18}       & \textbf{96.94±0.26}        \\ [2pt]
\bottomrule [1pt]
\end{tabular*}
\end{table*}

\begin{table*}
\centering
\caption{Class-Incremental Learning (F1-score/\%). Bold fonts highlight the optimal performance on each dataset in this experiment, while suboptimal results are underlined. ``\textbackslash" denotes the experiment that cannot be executed given the constraints of the task.}
\label{Table_6}
\begin{tabular*}{\textwidth}{p{2cm}<{\centering}|p{2cm}<{\centering}|p{2.2cm}<{\centering}p{2.2cm}<{\centering}p{2.2cm}<{\centering}|p{2.2cm}<{\centering}p{2.2cm}<{\centering}} 
\cline{1-3}\cline{3-5}\cline{5-7}
\toprule [1pt]
Model/Dataset & Task Type                & Cora             & CiteSeer         & PubMed           & Coauthor-CS      & Coauthor-Phy      \\ [2pt] 
\midrule [1pt]
MLP           & \multirow{11}{*}{\makecell[c] {Class-Incremental \\  \\ (Class-IR)}} & 55.16±1.47       & 59.32±1.52       & 77.96±1.96       & 85.24±1.08       & 86.83±1.66        \\[2pt] 
ChebNet       &                          & 76.32±0.67       & 67.63±1.77       & 79.46±1.53       & 87.49±0.74       & 90.13±0.52        \\[2pt] 
GCN           &                          & 79.28±0.55       & 69.17±0.63       & 84.62±0.48       & 89.23±0.68       & 91.44±0.37        \\[2pt] 
GraphSAGE     &                          & 80.53±0.44       & 71.20±0.84       & 83.88±0.23       & 88.84±0.54       & 90.78±0.48        \\[2pt] 
GAT           &                          & 82.86±0.38       & 69.73±0.92       & 83.26±0.43       & 89.77±0.26       & 91.85±0.16        \\[2pt] 
APPNP         &                          & 81.46±0.33       & 68.95±0.64       & 84.04±0.36       & 90.41±0.25       & 91.56±0.33        \\[2pt] 
FGNs          &                          & \underline{83.16±0.14}       & 73.10±0.32       & \underline{85.40±0.05}       & 91.11±0.18       & 92.04±0.32        \\[2pt] 
OTGNet          &                          & 83.04±0.20       & 73.14±0.25       & 85.12±0.33       & \underline{91.32±0.24}       & 92.05±0.16        \\[2pt] 
SEM          &                          & 81.18±0.24       & 72.41±0.65       & 84.06±0.16       & 89.92±0.20       & 91.25±0.22        \\[2pt] 
TACO          &                          & 82.26±0.32       & \underline{73.35±0.51}       & 84.42±0.24       & 90.95±0.27       & \underline{92.12±0.14}        \\[2pt] 
\textbf{BGML}          &                          & \textbf{83.33±0.12}       & \textbf{73.94±0.23}       & \textbf{85.53±0.08}       & \textbf{91.69±0.14}       & \textbf{92.38±0.25 }       \\[2pt] 
\bottomrule [1pt]
\end{tabular*}
\end{table*}

\subsection{Ablation Studies}
Table \ref{Table_7} presents the ablation research conducted on the BGML graph memory learning framework proposed in this article, focusing on the Cora dataset. To validate the correctness and effectiveness of the design concept presented in this paper, we conducted ablation experiments on two aspects: (1) Whether to adopt the MGHPL mechanism, utilizing second-level submodels to support weaker first-level submodels. (2) Whether to use the ISAO mechanism when introducing new incremental information, allowing them to make autonomous choices. Comparisons between model 2 and model 1, and model 4 and model 3 clearly demonstrate that the MGHPL mechanism, akin to brain-like cognition, can augment and rectify the shortcomings of coarse perception, leading to more accurate information cognition. Comparisons between model 3 and model 1, model 4 and model 2 demonstrate that allowing new information to select its ownership is the correct approach. Contrary to the prior weak information access, selecting the partition that best aligns with the information itself facilitates better integration into the cognitive system. Credible graph structures can avoid the impact of the old experience as much as possible. ISAO achieves the effect of alleviating the conflict between remembering and forgetting.
\begin{table}
	\centering
	\caption{Ablation studies. ``$ \surd  $" indicates that the component is included in the model, whereas ``$ \times  $" indicates that it is not included.}
	\label{Table_7}
		\begin{tabular}{c|cc|c}
			\toprule[1pt]
			\multicolumn{1}{c|}{\multirow{2}{*}{Model}} & \multicolumn{2}{c|}{Ablation}    & \multirow{2}{*}{F1-Score} \\ \cmidrule{2-3}
			\multicolumn{1}{c|}{}                       & \multicolumn{1}{c|}{Hierarchical Progressive} & \multicolumn{1}{c|}{Auto-attribution}                                           \\  \midrule [1pt]
			1  & $ \times  $ & $ \times $& 85.45±0.22\\
			2  & $ \surd  $  & $ \times $ & 86.07±0.24\\
			3  & $ \times $ & $ \surd $ & 85.93±0.18 \\
			4  & $ \surd $   & $  \surd$ & \textbf{86.30±0.15} \\
			\bottomrule[1pt]
	\end{tabular}
\end{table}
\subsection{Hyperparameters Analysis}\label{chap:5-G}
This section primarily examines the influence of hyperparameters on BGML memory performance. Hyperparameters are categorized into two groups, using the Cora dataset as an example for discussion. The first group includes hyperparameters related to graph partition in BGML, namely, the number of graph shards $k$ and $l$ at all levels. They are closely linked to the number of subsequent graph grains and sub-models at all levels. The second group comprises hyperparameters embedded in the model for retrieval: the number $\tau$ of sub-models that need to be supported retrieved by the progress-aware module, and the number of neighbors $\mu$ chosen after the new information independently selects shards.
\begin{figure}
	\setlength{\abovecaptionskip}{-0.5cm}
	\setlength{\belowcaptionskip}{0cm}
	\begin{center}
		\subfloat[Number of first-level graph shards]{
			\includegraphics[width=0.24\textwidth, height=2.8cm]{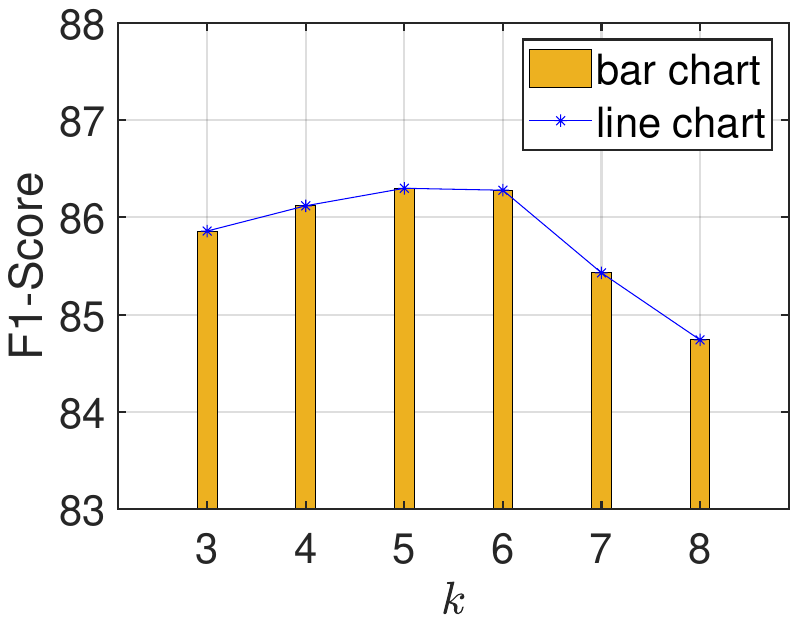}}
		\subfloat[Number of second-level graph shards]{
			\includegraphics[width=0.24\textwidth, height=2.8cm]{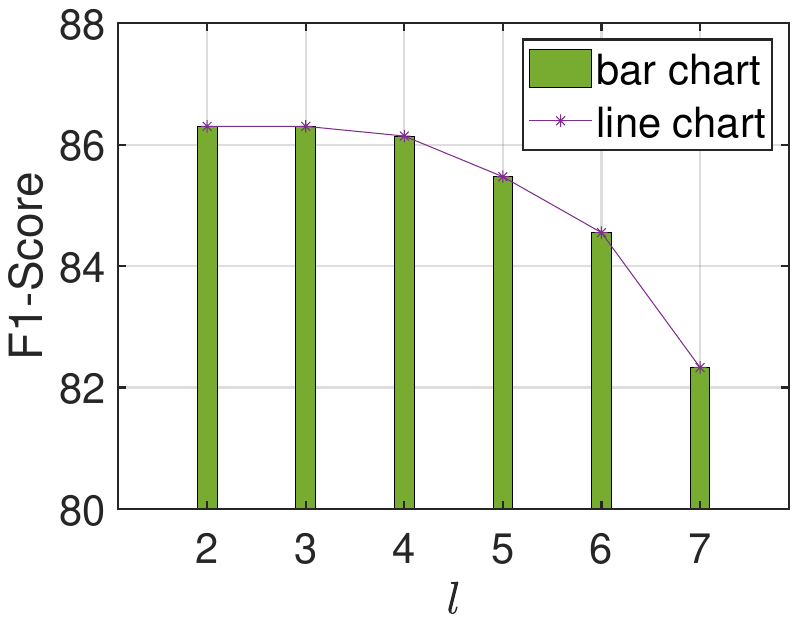}}
	\end{center}
        \vspace{0.35cm}
	\caption{The impact of $k$ and $l$.}\label{fig: k and l}
	
\end{figure}

\textbf{i) The influence of graph shard number hyperparameter ($k$ and $l$).} Fig. \ref{fig: k and l} (a) depicts the variation curve of the model's memory capacity across the empirical interval as the number of first-level graph shards $k$ changes. The entire curve exhibits an initial increase, followed by stabilization, and an eventual decline. The reason behind this phenomenon is quite evident: If the number of shards is too small, each submodel fails to capture enough node distinctions, resulting in insufficient decision-making information after posterior fusion. Conversely, an excessive number of shards leads to excessively scattered graph shards, resulting in insufficient data volume for the sub-model, thereby impeding complete model training. Fig. \ref{fig: k and l} (b) discusses the number of secondary graph shards $l$. By considering the support effect of the second-level submodel in the ablation experiment, we can conclude from this figure that a certain number of second-level submodels can effectively support the first-level submodel. However, excessively large or small numbers of second-level graph shards, as well as low-quality second-order submodels, may have counterproductive effects. Naturally, it becomes evident that for larger datasets, an appropriate increase in the number of corresponding graph shards is necessary.
\begin{figure}
	\setlength{\abovecaptionskip}{-0.5cm}
	\setlength{\belowcaptionskip}{0cm}
	\begin{center}
		\subfloat[Number of sub-models retrieved]{
			\includegraphics[width=0.24\textwidth, height=2.8cm]{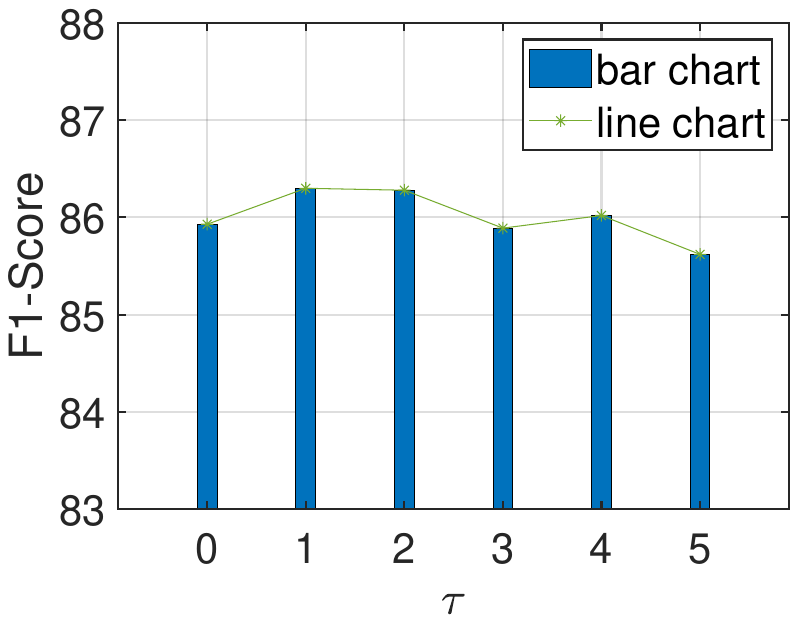}}
		\subfloat[Number of self-selected neighbors]{
			\includegraphics[width=0.24\textwidth, height=2.8cm]{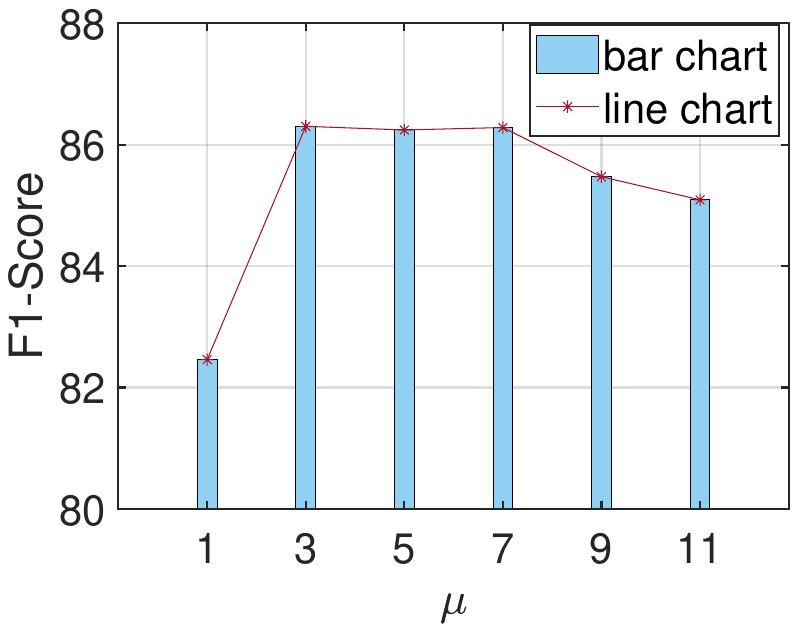}}
	\end{center}
        \vspace{0.35cm}
	\caption{The impact of $\tau$ and $\mu$.}\label{fig: tau and mu}
	
\end{figure}

\textbf{ii) The influence of retrieval hyperparameters in BGML ($\tau$ and $\mu$).} Fig. \ref{fig: tau and mu} (a) investigates the number $\tau$ of first-level sub-models in BGML that need support. Experimental results typically indicate that the model achieves optimal performance when only 1–2 models are supported. This suggests that BGML operates similarly to the brain mechanism, where learning first-level graph grains is usually sufficient to support decision-making in the model, except in select harder scenarios requiring further fine-grained perception. Fig. \ref{fig: tau and mu} (b) illustrates the impact of the number of node neighbors on model memory performance. Clearly, the model exhibits optimal performance when $\mu$ falls within a reasonable range. This phenomenon occurs because, after new information autonomously selects its ownership based on its information, a reasonable number of neighbors should be chosen among the shards for information dissemination and aggregation. A too-small number of neighbors will hinder the system's recognition of new information, while an excessive number will interfere with node learning due to excessive irrelevant information.

\subsection{Analysis of FR and IR Quantities}\label{chap:5-H}
\begin{figure}
	\setlength{\abovecaptionskip}{-0.5cm}
	\setlength{\belowcaptionskip}{0cm}
	\begin{center}
        \captionsetup[subfloat]{captionskip=0pt, justification=centering}
		\subfloat[Number of FR nodes \\ (with fixed 10 IR nodes)]{
            \includegraphics[width=0.16\textwidth, height=2cm]{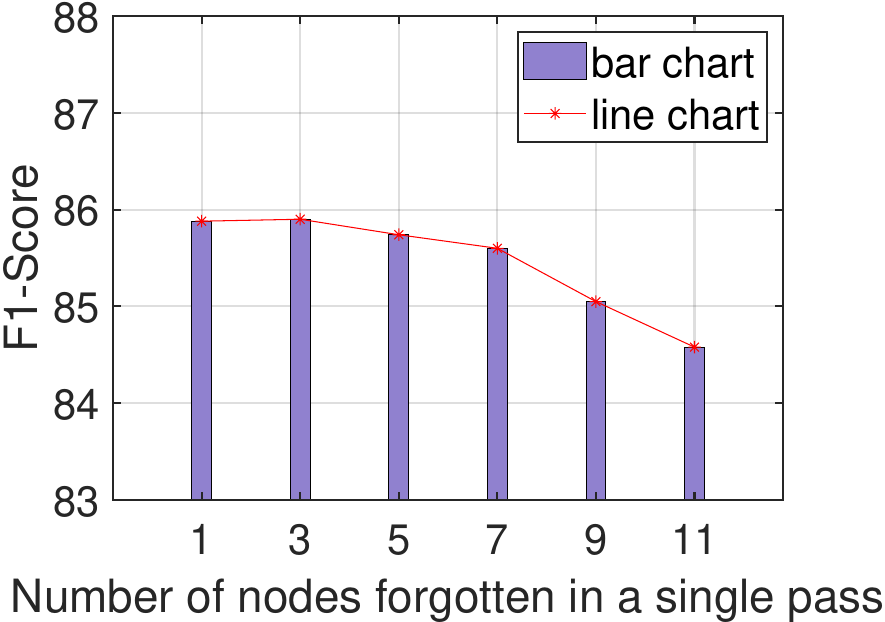}}
		\subfloat[Number of FR nodes \\ (with fixed 20 IR nodes)]{
			\includegraphics[width=0.16\textwidth, height=2cm]{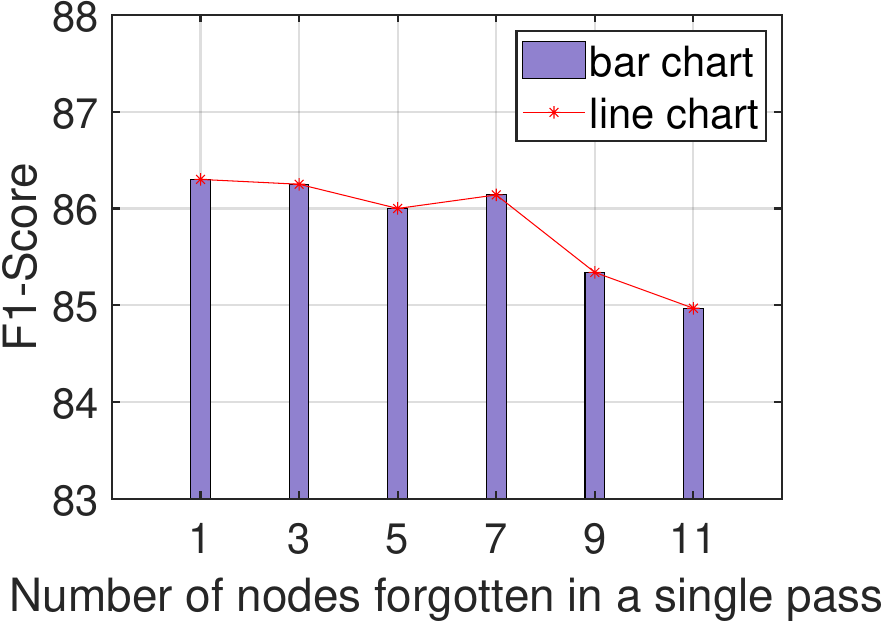}}
        \subfloat[Number of FR nodes \\ (with fixed 30 IR nodes)]{
			\includegraphics[width=0.16\textwidth, height=2cm]{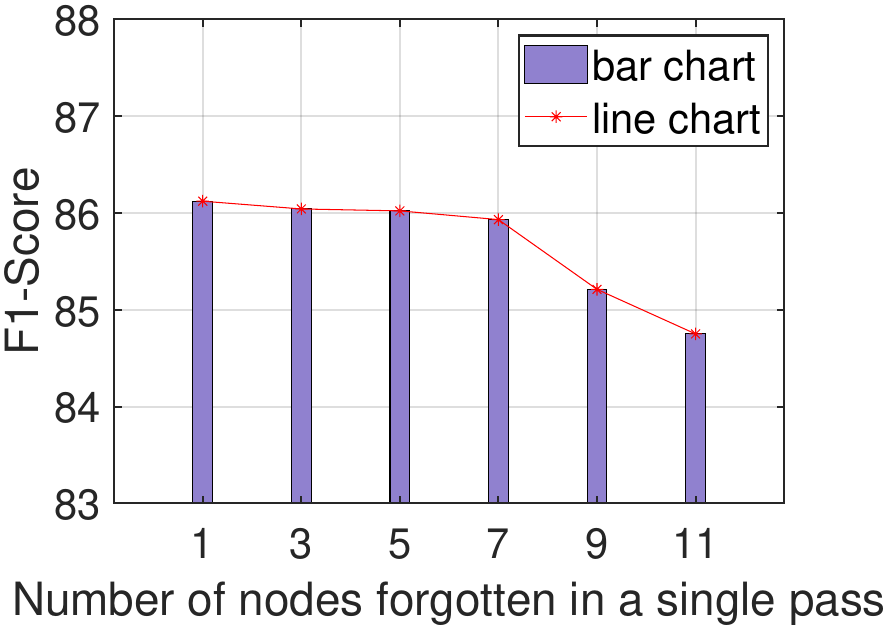}}
    \end{center}
    
    \vspace{-2em} 
    \begin{center}
        \captionsetup[subfloat]{captionskip=0pt, justification=centering}
		\subfloat[Number of IR nodes \\ (with fixed 1 FR nodes)]{
			\includegraphics[width=0.16\textwidth, height=2cm]{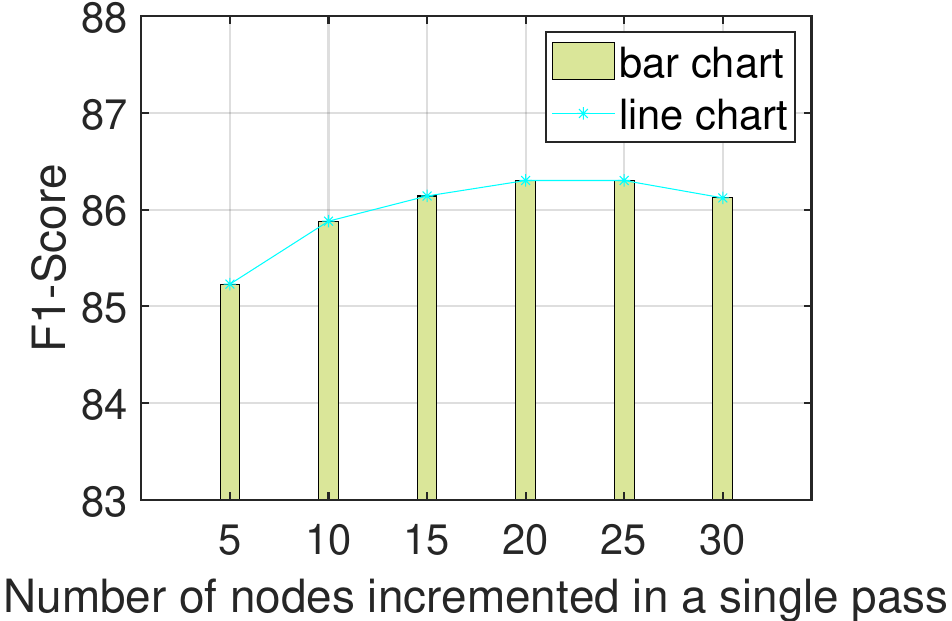}}
		\subfloat[Number of IR nodes \\ (with fixed 5 FR nodes)]{
			\includegraphics[width=0.16\textwidth, height=2cm]{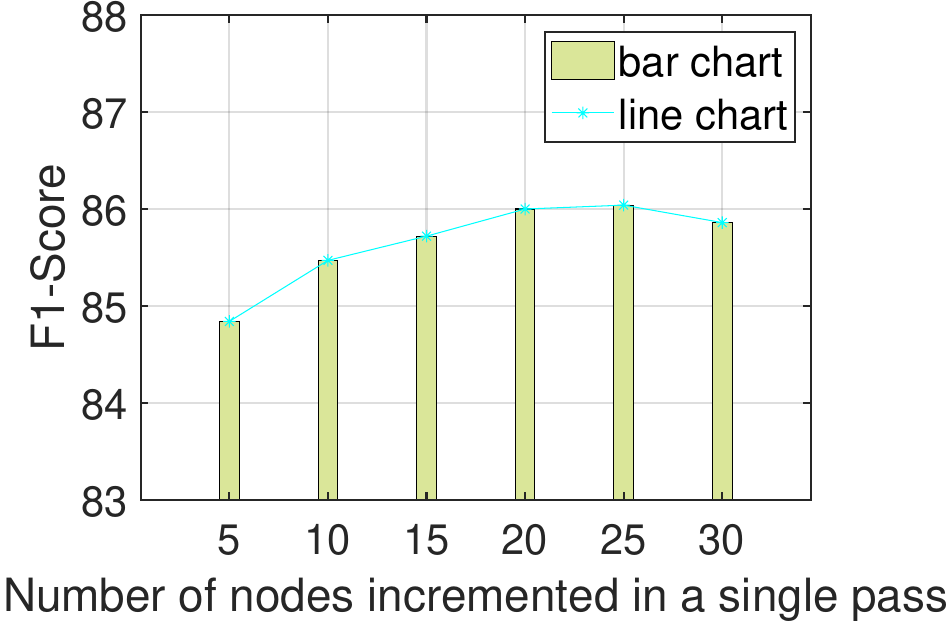}}
        \subfloat[Number of IR nodes \\(with fixed 9 FR nodes)]{
			\includegraphics[width=0.16\textwidth, height=2cm]{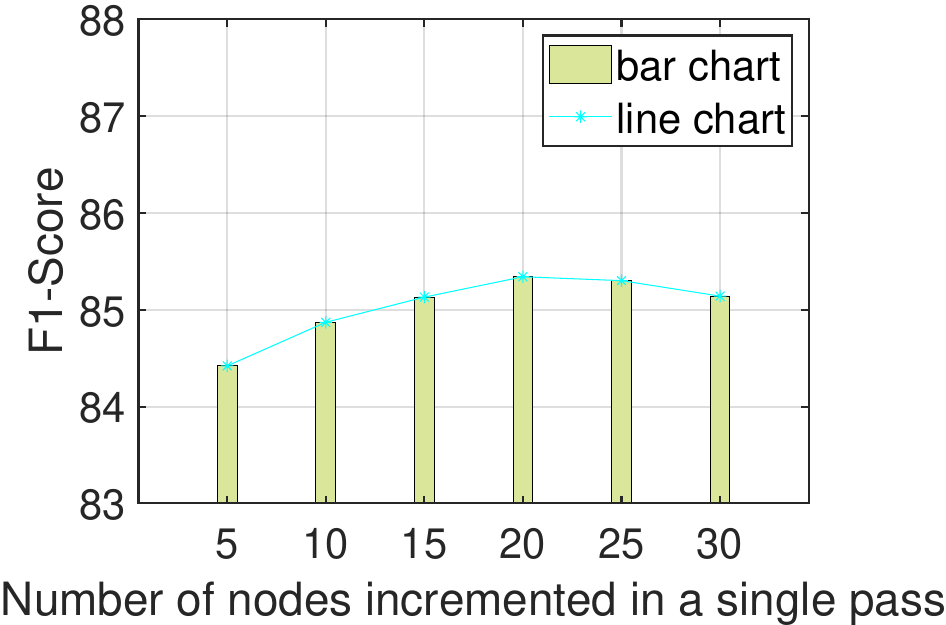}}
    \end{center}
 \vspace{0.35cm}
	\caption{The impact of node changes within a single timestamp in memory task.}\label{fig: change nodes}
	
\end{figure}
Fig. \ref{fig: change nodes} illustrates the effect of the number of FR and IR received by BGML within a single timestamp on the model's memory performance. The experimental results indicate that the model's performance remains relatively stable when processing a certain range of forgetting request instructions. However, if the model continuously receives a large number of FR over an extended period, the balance of nodes in each shard is disrupted. Additionally, after some time, the remaining nodes in certain shards may no longer be sufficient to support sub-model training. These factors directly contribute to the degradation of model performance. Furthermore, BGML can more calmly handle continuous streaming and incremental input. The experimental trend indicates that performance initially increases and then decreases as IR increase. This performance improvement stage occurs because continuous increments enable the model to absorb additional useful information. However, an excessive number of increments significantly damages the model at the previous moment, resulting in a decline in performance.

\section{Conclusion and Discussion}\label{chap:6}
In response to the challenge of continuous data influx and withdrawal in graph machine learning, this paper highlights the pressing issues of graph memory learning for the first time. Thus, inspired by the modular and hierarchical structure of the brain, as well as the brain network dynamics, the paper presents a general GML framework. BGML adeptly addresses diverse memory challenges in dynamic evolution scenarios and embodies the principle of selective remembering and forgetting. Experimentally, BGML outperforms other baseline models to varying degrees across different learning and memory tasks on the nine datasets.

\textbf{Limitation and Future Work.} This study uses nodes as instances study to investigate strategies for handling the influx and withdrawal of nodes in dynamic graphs. In practical scenarios, edge evolution is also prevalent. In the future, Our research will address edge-level tasks in GML problems. Additionally, during the course of this research, potential directions for future exploration emerged, particularly in mitigating the conflict between memory and forgetting. Specifically, \emph{(1) whether the experiences of individual sub-models in processing evolution requests for similar structures can be stored and leveraged. there exist interareal communication subspaces in the brain, which enhance learning. How can we establish connections between BGML's sub-models without compromising its efficient ability? (2) The conflicts stem from the need to distinguish between memory and forgetting requests. BGML sequentially processes FR and IR within the same time stamp. Is it possible to cleverly overwrite or modify the information slated for forgetting using IR, thereby offering a more uniform approach to handling memory issues?}

\bibliography{IEEEabrv, mybibfile}
\bibliographystyle{IEEEtran}

\begin{IEEEbiography}[{\includegraphics[width=1in,height=1.25in,clip,keepaspectratio]{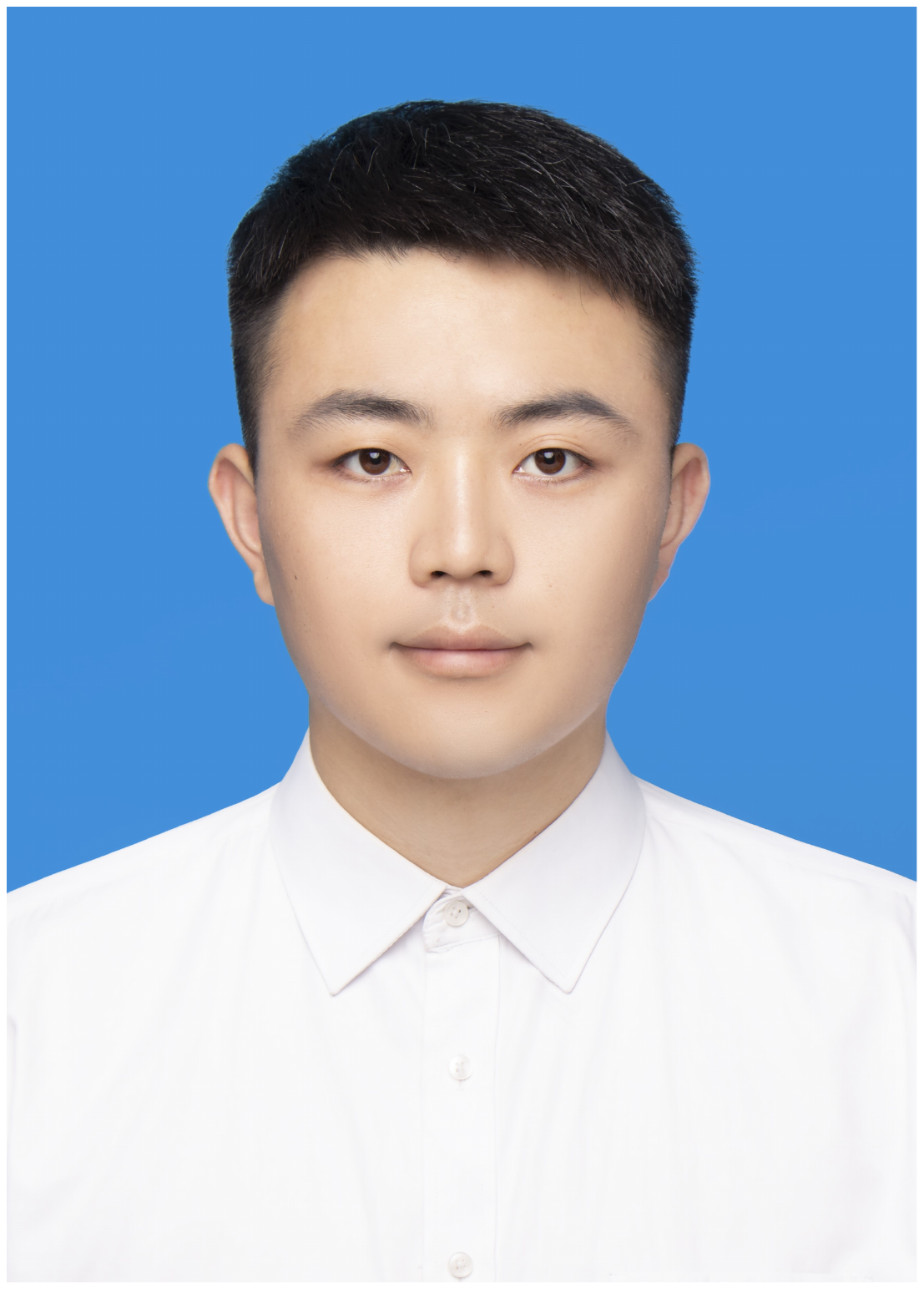}}]{Jiaxing Miao} received his M.S. degree from the School of Sciences, China Jiliang University, Hangzhou, China, in 2023. He is currently pursuing a Ph.D. degree from the College of Electronics and Information Engineering, Tongji University, Shanghai, China. His current research interests mainly include graph representation learning, lifelong learning, and machine unlearning.
\end{IEEEbiography}


\begin{IEEEbiography}[{\includegraphics[width=1in,height=1.25in,clip,keepaspectratio]{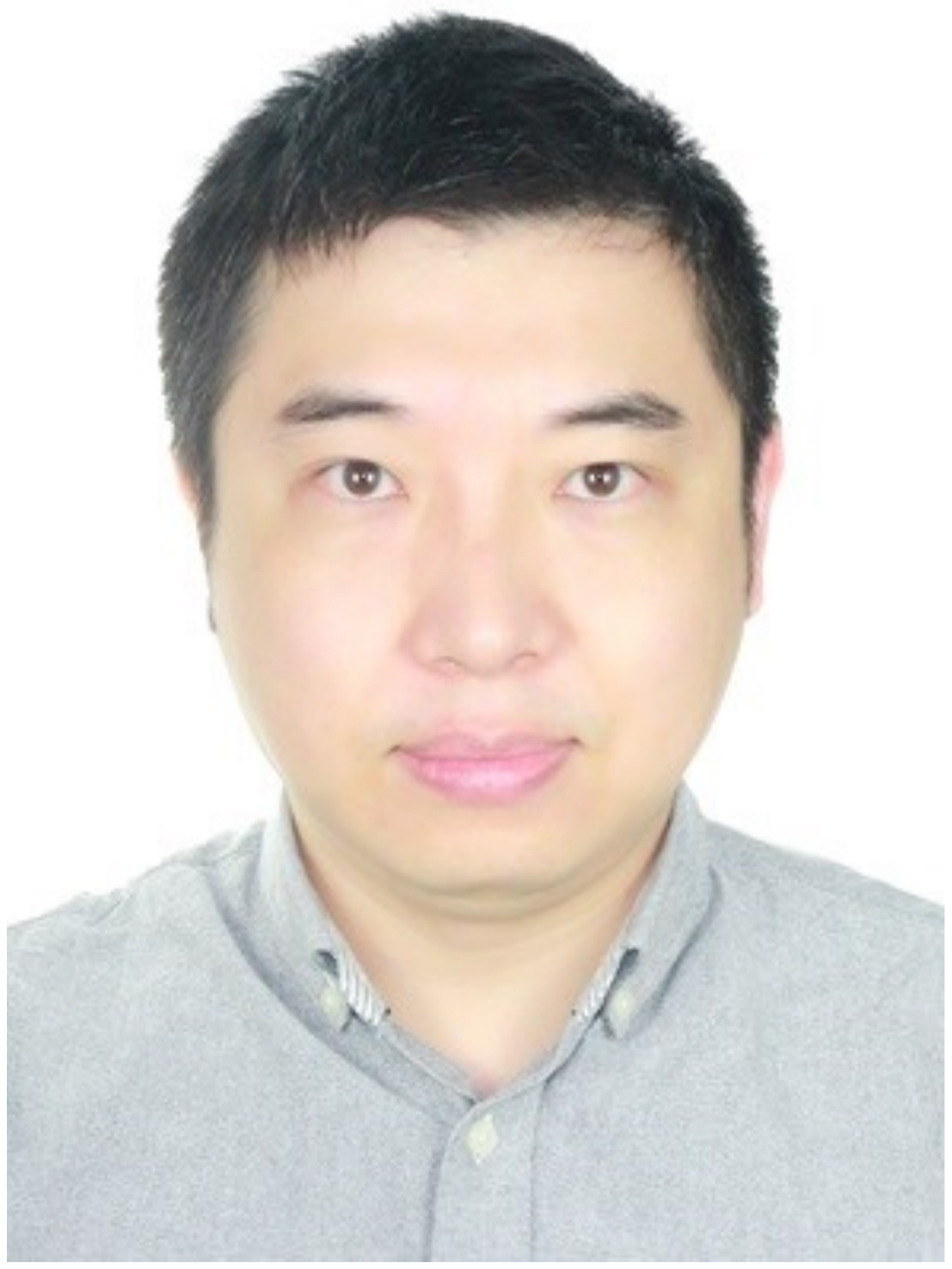}}]{Liang Hu} (@M in 2022) is a full professor with Tongji University.  His research interests include recommender systems, machine learning, data science and interdisciplinary intelligence. He has published a number of papers in top-rank international conferences and journals, including WWW, IJCAI, AAAI, ICDM, NeurIPS, TOIS, TNNLS, TKDE. He has been invited as the program committee of more than 50 top-rank AI international conferences, including AAAI, IJCAI, ICDM, CIKM, CVPR, and KDD. 
\end{IEEEbiography}


\begin{IEEEbiography}[{\includegraphics[width=1in,height=1.25in,clip,keepaspectratio]{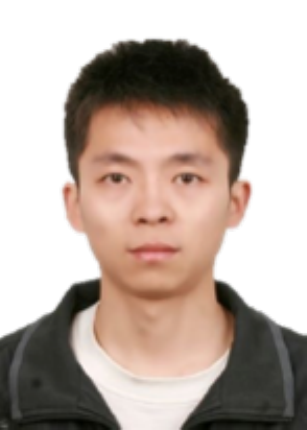}}]{Qi Zhang}
received the PhD degree from the Beijing Institute of Technology, Beijing, China, and the University of Technology Sydney, Sydney, NSW, Australia, in 2020.
He is currently a Research Fellow at Tongji University, Shanghai, China. He has authored high-quality papers in premier conferences and journals, including AAAI, IJCAI, SIGIR, TKDE, TNNLS, and TOIS. His research interests include collaborative filtering, sequential recommendation, learning to hash, and MTS analysis.
\end{IEEEbiography}


\begin{IEEEbiography}[{\includegraphics[width=1in,height=1.25in,clip,keepaspectratio]{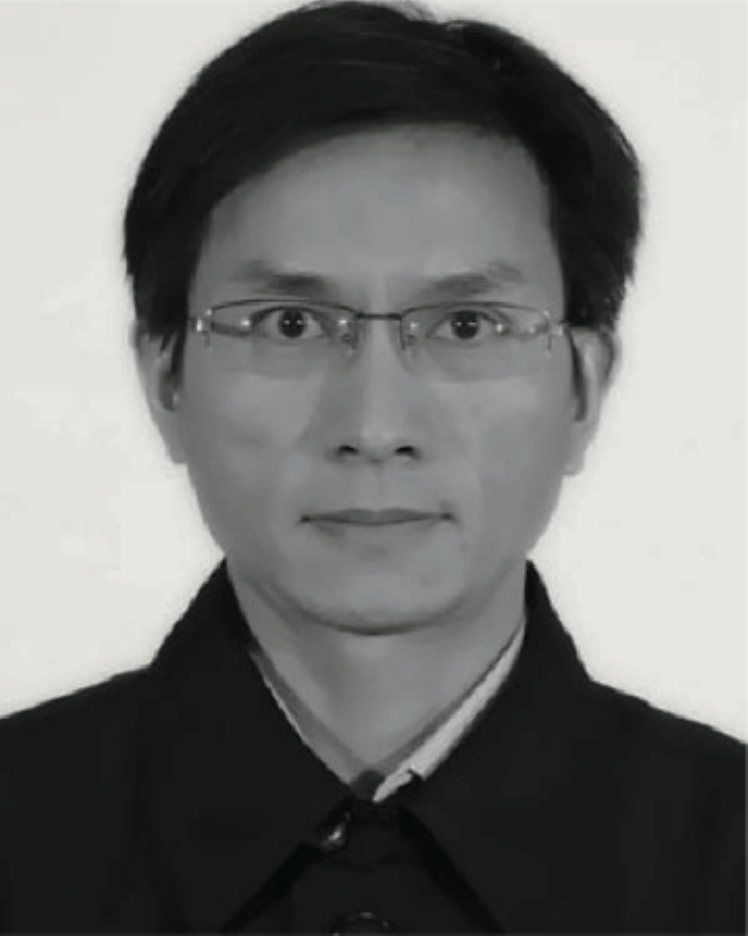}}]{Longbing Cao}
(@SM in 2006) received the PhD degree in pattern recognition and intelligent systems from the Chinese Academy of Science, China, and the PhD degree in computing sciences from the University of Technology Sydney, Australia. He is a distinguished chair professor with Macquarie University, an ARC future fellow (professorial level), and the EiCs of IEEE Intelligent Systems and J. Data Science and Analytics. His research interests include AI, data science, machine learning, behavior informatics, and their enterprise applications.
\end{IEEEbiography}

\appendices


\vfill
\end{document}